%% file: main.tex
\documentclass[conference]{IEEEtran}
\usepackage{times}

\usepackage[numbers]{natbib}
\usepackage{multicol}
\usepackage[bookmarks=true]{hyperref}

\usepackage{microtype}
\usepackage{amsmath,amssymb}
\usepackage{paralist}
\usepackage{mdframed}
\usepackage{xcolor}
\usepackage{tikz}
\usepackage{amsthm}
\usepackage{amssymb}
\usepackage{nicefrac}
\usepackage{amsthm}
\usepackage{colonequals}

\newtheorem{proposition}{Proposition}
\theoremstyle{remark}
\newtheorem{remark}{Remark}
\newtheorem{example}{Example}

\usepackage{import}
\usepackage{pgf}

\usepackage{todonotes}
\usepackage{pgfplots}
\usepackage{algorithm, algorithmicx}
\usepackage[noend]{algpseudocode}

\usepgfplotslibrary{fillbetween}
\usetikzlibrary{patterns,arrows,backgrounds,calc,shapes,shadows,decorations.pathmorphing,decorations.pathreplacing,automata,shapes.multipart,positioning,shapes.geometric,fit,circuits,trees,shapes.gates.logic.US,fit,decorations.markings}

\tikzset{sstate/.style={circle, draw=black, inner sep=1pt,minimum height=6mm}}
\tikzset{astate/.style={diamond, draw=black, inner sep=1pt}}
\tikzset{tstate/.style={rectangle, draw=black, inner sep=1pt, minimum height=5mm}}
\tikzset{actnode/.style={fill=black, inner sep=1pt}}
\tikzset{elab/.style={auto,font={\fontsize{9pt}{12}\selectfont}}}

\tikzset{cross/.style={cross out, draw=black, minimum size=2*(#1-\pgflinewidth), inner sep=0pt, outer sep=0pt},
cross/.default={1pt}}

\makeatletter
\let\MYcaption\@makecaption
\makeatother

\usepackage[font=footnotesize]{subcaption}

\makeatletter
\let\@makecaption\MYcaption
\makeatother

\newcommand{\Nat}{\mathbb{N}}
\newcommand{\NN}{\mathbb{N}}

\newcommand{\sg}{\mathcal{G}}
\renewcommand{\path}{\xi}

\newcommand{\sched}{\sigma}

\newcommand{\Distr}{\ensuremath{\textsl{Distr}}}
\newcommand{\act}{\alpha}

\newcommand{\Act}{A}
\newcommand{\scp}{p}
\newcommand{\scthreshold}{\mathbf{p}}

\newcommand{\target}{s_{\top}} 
\newcommand{\sink}{s_{\bot}}
\newcommand{\rndp}{h}
\newcommand{\randomness}{\mathbf{h}}

\newcommand{\last}[1]{\mathsf{last}({#1})}
\newcommand{\pOneSched}{{\sigma_\pOne}}
\newcommand{\pOneSchedPrime}{{\sigma'_\pOne}}

\newcommand{\pTwoSched}{{\sigma_\pTwo}}

\newcommand{\rat}{\lambda}
\newcommand{\pOne}{\mathsf{ego}}
\newcommand{\pTwo}{\mathsf{env}}
\newcommand{\player}{\mathsf{i}}

\newcommand{\solutions}{\mathbb{S}}
\newcommand{\EnAct}{\Act}
\newcommand{\solfuncp}{f_\mathbb{S}}
\newcommand{\scopt}{\scp^{*}}
\newcommand{\rndopt}{\rndp^{*}}
\newcommand{\pareto}[1]{\mathcal{F}_{#1}}
\newcommand{\smoothmax}{\operatorname*{\mathrm{smax}}}
\newcommand{\indicator}[1]{[#1]}
\newcommand{\Succ}{\mathsf{Succ}}

\newcommand{\supp}{\mathsf{support}}
\newcommand{\ppthreshold}{\mathbf{d}}

\newcommand{\scmin}{\scp^{-}}
\newcommand{\rndmin}{\rndp^{-}}
\newcommand{\pathslbl}{\mathsf{Paths}}
\newcommand{\Paths}[2][]{\pathslbl^{#2}_{#1}}
\newcommand{\POnePaths}[2][]{{[\pathslbl^{#2}_{#1}]_\pOne}}
\newcommand{\PTwoPaths}[2][]{{[\pathslbl^{#2}_{#1}]_\pTwo}}
\newcommand{\PlayerPaths}[2][]{{[\pathslbl^{#2}_{#1}]}_{\player}}

\newcommand{\causalprob}[2]{\Pr(#1\mid\mid#2)}

\newcommand{\expOver}[2]{\mathbb{E}_{#1}[#2]}
\newcommand{\soft}{\varphi}
\newcommand{\hard}{\psi}
\newcommand{\rv}[1]{{\mathcal{#1}}}  % Random Variable
\newcommand{\droneEnv}{{D_{\text{new}}}}
\newcommand{\droneEgo}{{D_{\text{test}}}}

\newcommand{\eqdef}{\mathrel{\stackrel{\makebox[0pt]{\mbox{\normalfont\tiny def}}}{=}}}
\newcommand{\mypara}[1]{\smallskip\noindent{\bf #1.}}
\newcommand{\propref}[1]{{Prop.~\ref{prop:#1}}}

\begin{document}

\title{Entropy-Guided Control Improvisation}

% You will get a Paper-ID when submitting a pdf file to the conference system
% \author{Author Names Omitted for Anonymous Review. Paper-ID 177}

\author{
  \authorblockN{
    Marcell Vazquez-Chanlatte\authorrefmark{1},
    Sebastian Junges\authorrefmark{1},
    Daniel J. Fremont\authorrefmark{2},
    Sanjit A. Seshia\authorrefmark{1}
  }
  \authorblockA{
    University of California, \{Berkeley\authorrefmark{1}, Santa Cruz\authorrefmark{2}\}
  }
}

% avoiding spaces at the end of the author lines is not a problem with
% conference papers because we don't use \thanks or \IEEEmembership

%Homer Simpson\authorrefmark{2},
%Montgomery Scott\authorrefmark{3} and
%Eldon Tyrell\authorrefmark{4}}
%Telephone: (800) 555--1212, Fax: (888) 555--1212}
%\authorblockA{\authorrefmark{4}Tyrell Inc., 123 Replicant Street, Los Angeles, California 90210--4321}}

\maketitle

\begin{abstract}
  % Sentence 1: State the problem
  High level declarative constraints provide a powerful (and popular)
  way to define and construct control policies; however, most
  synthesis algorithms do not support specifying the degree of
  randomness (unpredictability) of the resulting controller.
  % Sentence 2: State the consequences
  In many contexts, e.g., patrolling, testing, behavior prediction,
  and planning on idealized models, predictable or biased controllers
  are undesirable.
  % Sentence 3: State your solution
  To address these concerns, we introduce the \emph{Entropic Reactive
    Control Improvisation} (ERCI) framework and algorithm which
  supports synthesizing control policies for stochastic games that are
  declaratively specified by (i) a \emph{hard constraint} specifying
  what must occur, (ii) a \emph{soft constraint} specifying what
  typically occurs, and (iii) a \emph{randomization constraint}
  specifying the unpredictability and variety of the
  controller, as quantified using causal entropy.
  % Sentence 4: State the consequences of the solution
  This framework, extends the state of the art by supporting
  arbitrary combinations of adversarial and probabilistic uncertainty
  in the environment. ERCI enables a flexible modeling formalism which
  we argue, theoretically and empirically, remains tractable.
\end{abstract}

\IEEEpeerreviewmaketitle

%\maketitle\sj{Daniel?}
%\begin{abstract}
%	Efficacious controller synthesis is a key ingredient in the design and analysis of complex systems. We study the design of controllers that have a high entropy, that is, whose behavior or nature is surprising. The synthesis of such controllers is key in domains like testing and security. 
%	In particular, our paper studies control improvisation and compares them with randomly sampling adequate policies. The only difference in obtained policies is in their notion of entropy, but the problems are significantly different.  We illustrate and contrast their merits and limitations. Furthermore, we provide algorithms that solve both control improvisation problems. Prominently, we solve the control improvisation problem for Markov decision processes by relating it to recent results from inference from demonstrations, and then extend this approach to stochastic games. We present a prototypical implementation that efficiently solves controller synthesis problems from the security and testing domain. 
%\end{abstract}
\section{Introduction}
\input{introduction}
\section{Motivating Example}
\label{sec:motivating}

\input{example}

\section{Problem Statement}
\label{sec:problem}
This section formalizes the novel Entropic Reactive Control
Improvisation (ERCI) problem.  We start with some necessary
definitions and notations on stochastic games.

\input{problem}

\input{aspects}

\input{mdp}

\section{The Control Improvisation Problem for SGs}\label{sec:sgs}
\input{sg}
%\begin{mdframed}
%
%\begin{compactenum}
%	\item $\Pr^\sg_{\langle \sched_1,\sched_2 \rangle}(\eventually{h} T) \geq 1$
%	\item $\Pr^\sg_{\langle \sched_1,\sched_2 \rangle}(\eventually{h} G) \geq \lambda$ 
%\end{compactenum}
%\end{mdframed}
%\sj{Define randomly selected policy}
%\sj{Describe in terms of pMDPp}

\section{Implementation and Empirical Evaluation}
\label{sec:empirical}
\input{experiment.tex}

\input{related}
\input{conclusion}

{\vspace{0.5em} 
  \noindent\textbf{Acknowledgments}:
This work is partially supported by NSF grants 1545126 (VeHICaL), 1646208 and 1837132, by the DARPA contracts FA8750-18-C-0101 (Assured Autonomy) and FA8750-20-C-0156 (SDCPS), by Berkeley Deep Drive, and by Toyota under the iCyPhy center.}

\bibliographystyle{plainnat}
\bibliography{bibliography}

\input{proofs}

\end{document}

%% file: introduction.tex
% Declarative Constraints ar neat idea.
The use of declarative specifications, e.g. in the form of temporal logic formulas, has become a popular way to construct high-level robot controllers~\cite{DBLP:conf/iros/HorowitzWM14, DBLP:conf/rss/WongEK14, DBLP:conf/iros/HeLKV17, DBLP:conf/icra/FuATP16, DBLP:conf/icra/HeWKV19, DBLP:journals/arobots/MoarrefK20, DBLP:conf/icra/KantarosM0P20}.
% Synthesis closes the gap.
Given a user provided specification, \emph{synthesis} algorithms aim
to automatically create a control policy that ensures that the
specification is met, or explain why such a policy does not
exist. Together, synthesis and declarative specifications facilitate
quickly and intuitively solving a wide variety of control tasks.  For
example, consider a delivery drone operating in a workspace. One may
specify the drone should ``within 10 minutes, visit four locations (in any
order) \emph{and} avoid crashing.''. A synthesis tool may then create a
finite state controller which guarantees this specification is met,
under a particular world model.
% Declarative Synthesis need not produce variety.  
Importantly, while many controllers may conform to the
provided specification, many synthesis algorithms provide a
single, often deterministic, policy.  For instance, in our drone
example, a synthesized controller may generate only a single path
through the workspace.

% On the importance of being varied.
In some settings, such policies are undesirable.  First, in
many tasks, the predictability (or bias) of the policy may be a
liability.  Examples include
patrolling~\cite{DBLP:journals/ior/AlpernMP11}, behavior prediction
and inference~\cite{DBLP:conf/cav/Vazquez-Chanlatte20}, and creating
controller harnesses for fuzz testing (see motivating
example in Sec.~\ref{sec:motivating}). Second, synthesis algorithms work on \emph{idealized}
models, and thus any policy that overcommits to any given model quirk
may in practice yield poor performance. In such settings,
randomization is known to make policies more robust against worst-case
deviations~\cite{mceThesis, maxEntAnswer}. Unfortunately, classical 
synthesis methods result in policies that need not (and typically do
not) exhibit randomization.

% Propose CI and highlight new features.
To address these potential deficits, we advocate for the adoption of
the recently proposed \emph{control
improvisation}~\cite{DBLP:conf/cav/FremontS18,DBLP:conf/fsttcs/FremontDSW15}
framework, in which one specifies a controller with three types of
declarative constraints. (i) \emph{Hard constraints} that, as in the
classical setting, must hold on every execution, (ii) \emph{soft constraints} that should hold 
on most executions, and (iii) \emph{randomization
constraints} that ensure that a synthesized policy does not overcommit to a particular action or behavior. 
The key challenge when solving control improvisation is that randomization and performance, in the form of soft constraints, constitute a natural trade-off.

So far, control improvisation has  only be studied in 
nondeterministic domains where uncertainty is resolved
adversarially~\cite{DBLP:conf/cav/FremontS18}. This assumption is often too restrictive and leads
(together with the soft/hard constraints) to conservative policies or
common situations in which the synthesis algorithm cannot be employed
at all. To overcome this weakness, we develop a theory of control
improvisation in stochastic games which admit
\emph{arbitrary combinations of nondeterministic and probabilistic
uncertainty}, including unknown or imprecise transition
probabilities. 

Technically, we formulate our problem on \emph{simple stochastic
games}~\cite{DBLP:conf/dimacs/Condon90}, an extension of \emph{Markov decision processes} (MDPs) that divides states
between controllable states and uncontrollable (or adversarially
controlled) states. \emph{Soft constraints} are finite horizon
temporal properties with a threshold on the worst-case probability of
the property holding by the end of the episode. \emph{Hard
constraints} are soft constraints to be satisfied with probability 1. In
contrast to other work on control improvisation, we adopt causal entropy as a natural means to formalize \emph{randomness
constraints}.  Causal entropy is a prominent notion in directed
information theory~\cite{DirectedInfoTheoery} that strongly correlates with robustness in the
(inverse) reinforcement learning setting~\cite{mceThesis,
maxEntAnswer}. We refer to this variant of control improvisation as
\emph{Entropic Reactive Control Improvisation} (ERCI) and show that ERCI
conservatively extends reactive control improvisation~\cite{DBLP:conf/cav/FremontS18} to stochastic
games. More precisely, entropy can
be used in the non-stochastic setting and yields results analogous to
reactive control improvisation. ERCI also extends  classical policy synthesis in stochastic games, i.e. synthesis in absence of randomness constraints as, e.g., implemented in PRISM-games~\cite{DBLP:journals/sttt/KwiatkowskaPW18}.

%We argue that soft constraints can naturally be considered as an
%optimization objective which one can trade-off for more randomization.
%Indeed, our method strongly relies on the computation of a
%Pareto front that explores the trade-off between randomization and
%optimizing the soft constraint using the notion of rationality. This
%means that rather than asking the user to fix rather arbitrary
%threshold values for both types of constraints, we may visualize the
%trade-off between these two entities.
%
\mypara{Contributions}
In summary, this paper contributes ERCI, an algorithmic way to trade
performance and randomization in stochastic games. As we motivate in
the example below, games that combine both adversarial and
probabilistic behavior in an environment allow for modeling
flexibility, facilitating applicability to new domains. To support this
extension, the paper proposes and shows the benefits of formulating
randomization constraints with causal entropy.  Finally, this work
contributes the necessary technical machinery and a prototype 
implementation. Combined, our theoretical and empirical analysis
suggest that the ERCI framework contributes a tractable and flexible
modeling formalism.

\mypara{Overview} This paper is structured as follows. We begin with a
motivating example (Sec.~\ref{sec:motivating}). Then we provide
preliminaries and formalize the ERCI problem statement
(Sec.~\ref{sec:problem}). Next, we cast ERCI
as a multi-objective optimization problem and study properties of the
solution set (Sec.~\ref{sec:convex}). With this technical machinery developed,
Sec.~\ref{sec:mdps} re-frames existing literature on maximum causal
entropy inference and control to derive an algorithm for MDPs.  Then in Sec.~\ref{sec:sgs}, we provide an
algorithm for the general case of stochastic games. We conclude with
an empirical evaluation (Sec.~\ref{sec:empirical}) and a comparison
with related work, e.g., other control improvisation
formulations (Sec.~\ref{sec:related}). Proofs are attached in Sec.~\ref{sec:proofs}.

%%% Local Variables:
%%% mode: latex
%%% TeX-master: "main"
%%% End:

%% file: example.tex
\begin{figure}
  \centering \scalebox{0.33}{
    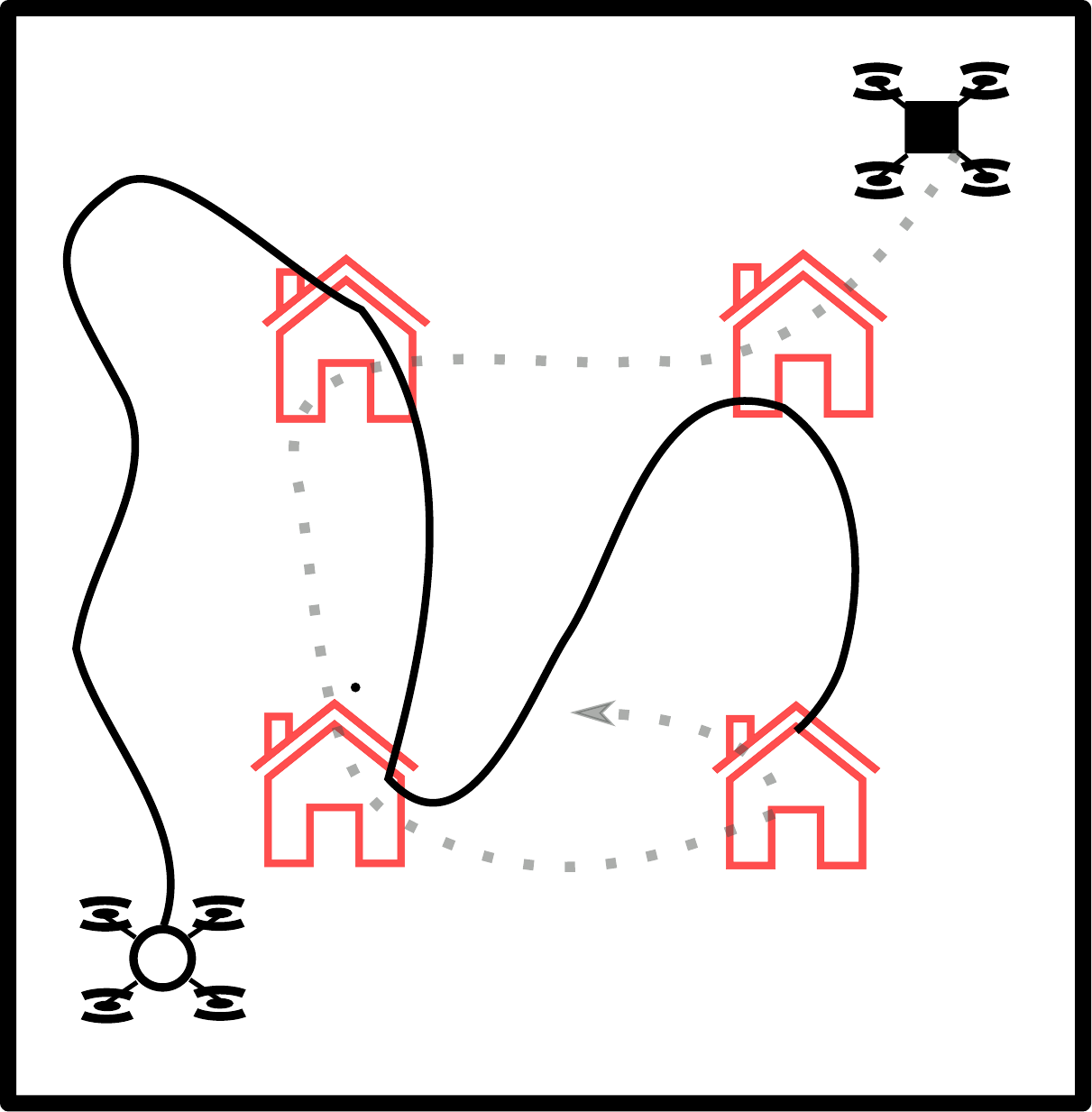 }
  \caption{ Illustration of delivery drone testing example. The goal
    is to synthesize a policy for the bottom left (white circle) drone
    to test the controller of the top right (black square) drone. Ideally,
    the synthesized policy should be as randomized as possible to avoid
    testing bias.\label{fig:motivating} }
\end{figure}

% Setup Testing Premise.
We consider a scenario in which a regulatory agency
wishes to certify the safety and performance of a new delivery drone $\droneEnv$.
As part of the process, the agency runs $\droneEnv$ through a series of tests. For example, given a certain
delivery route, the agency investigates whether $\droneEnv$ successfully delivers packages while
avoiding \emph{other} delivery drones. To execute this test, the agency decides to
synthesize a controller for another delivery drone, $\droneEgo$, to test if
$\droneEnv$ can be certified.

% Paint a picture.
Concretely,
% suppose we have high-level models (e.g., motion planners)
%for $\droneEnv$ and $\droneEgo$, 
suppose we command
$\droneEnv$ to continuously visit four houses in some workspace. We
illustrate such a scenario in~Fig.~\ref{fig:motivating}, in which
$\droneEnv$ and $\droneEgo$ are shown as black square and white circle
drones respectively.  For this test scenario, the regulatory agency,
wishes to exam how $\droneEnv$ responds to delivering packages to the
red houses in the presence of $\droneEgo$. In particular, it would like to let $\droneEgo$ also deliver packages while avoiding
$\droneEnv$. Importantly,
to properly exercise $\droneEnv$, $\droneEgo$ should show a \emph{variety} of behaviors meeting  the specification, and the behaviors should not be biased to any behavior beyond the given specification.
% Ideally, this $\droneEgo$'s policy should be as
%un-biased as possible to exercise $\droneEnv$ on a variety of
%scenarios, and ideally capture and sensor, software, or hardware
%errors, e.g., a bug in a machine learned perception sensor.

% Frame as RCI.
With the ERCI framework, the agency may formalize the above scenario with the
following constraints on $\droneEgo$:
\begin{enumerate}
\item (\emph{hard constraint}) Ensure that the two drones \emph{never} collide.
\item (\emph{soft constraint}) With probability at least $.8$, visit all four houses within 10 minutes.
\item (\emph{randomness constraint}) Perform this task as unpredictably as possible.
\end{enumerate}
What  remains is to synthesize a controller given the constraints
\emph{and} the world model. At this point, it is worth examining more
closely how one models $\droneEnv$'s controller when synthesizing
$\droneEgo$. We illustrate by examining three models. In all models, we capture the behaviors of $\droneEnv$ and $\droneEgo$. We focus on $\droneEnv$, but the ideas carry over to modeling the actuation of $\droneEgo$.
%
%\mypara{Deterministic Model}
%In the simplest case, the manufacture might guarantee that a
%$\droneEgo$ will deliver packages on a fixed route.\sj{But it also needs to respond to other drones, so I do not find this fixed route particularly convincing. } While tempting to
%believe, such a route may be hard to a-priori compute and sensitive to
%the exact details of the test workspace.

\mypara{Nondeterministic Model}
The simplest approach to modeling is not to make any assumptions about  $\droneEnv$ beyond what already has been established. Here, we model that the houses are visited either in clockwise or counter-clockwise order but that it may switch direction at \emph{any time}. 
Such a model is too liberal and our assumptions under which we plan the behavior for $\droneEgo$ is too pessimistic, which leads to a bad test set.
First, if $\droneEnv$ is unrestricted, then $\droneEgo$'s behavior is severely limited, as it must behave conservatively to avoid collisions under all possible motions by $\droneEnv$ (even very unlikely motions). This limitation restricts the variance of its behavior, and it will not test $\droneEnv$'s true behavior. 
A purely non-deterministic model for $\droneEnv$ thus may not lead to the synthesis of adequate behavior for $\droneEgo$. 
 %Such a model is clearly too pessimistic -- there is no policy that satisfies
%our constraints -- and frankly unrealistic within the context.
%Furthermore, even if there existed a winning policy, such a pessimistic
%model would result in an unnecessarily conservative (and thus predictable)
%test controller, limiting its utility.

\mypara{Stochastic Model}
Rather than the pessimistic nondeterministic (or adversarial)  assumption, we may collect data about $\droneEnv$ and construct a stochastic model, e.g., using inverse reinforcement learning~\cite{DBLP:conf/icml/NgR00}.
Concretely (but simplified), after examining the data, one observes that $\droneEnv$ appears
to flip a biased coin with fixed probability $p$ whenever it reaches a house to decide whether or
not to turn around. 
This models $\droneEnv$ much more precisely, and allows for more targeted test by $\droneEgo$.

%The resulting Markov Decision Process may be
%sufficiently correct to synthesize adequate improvisers without
%being too pessimistic.

\mypara{Nondeterministic and Stochastic Model} However, a natural criticism for stochastic
models is the dependence on \emph{fixed} probabilities. 
Obtaining such probabilities with confidence requires many tests which defeat the purpose of our test setup, and making point-estimates from little data may not create faithful models of the actual behavior.
% In our
%example, we may have observed $\droneEnv$'s behavior and extracted
%(point-)estimate probabilities, but these probabilities may still have
%non-trivial error margins or could be sensitive to changes in the
%workspace.
  In absence of enough (or reliable) data, we can arbitrarily combine
nondeterministic choices and stochastic behavior. We may use stochastic abstractions for parts that we can faithfully model, and nondeterministic behavior in absence of data.
In particular, we
support interval-valued transition probabilities.  Consider the
delivery-drone $\droneEnv$. Rather than inferring a point-estimate
from data, we may have inferred that the probability of turning around
is in the interval $[p - \varepsilon, p + \varepsilon]$ for adequate
values of $p$ and $\varepsilon$.  Furthermore the actual probability
may even depend on aspects of the current state.

% Bring it all together.
\mypara{ERCI as a unifying framework}
The strength of the (entropy-guided) control improvisation framework
is that we can combine all these aspects into a single and thus flexible 
computational model. 
In particular, the models above are captured by a 2-player game, a 1.5-player game (MDP) and a 2.5-player game (stochastic game, SG), respectively.
In all cases, the first player controls the behavior of $\droneEgo$ and this controller is to be synthesized. 
We contribute an algorithm that synthesizes a controller  that
 maximally randomizes in all of the formalisms discussed above. In the
coming sections, we shall formally define the ERCI problem, highlight
that there is an implicit trade-off between performance of the soft
constraint and unpredictability, and provide an algorithm solving ERCI
for~SGs.

%% file: imgs/motivating_example.pdf_tex
%% Creator: Inkscape inkscape 0.92.5, www.inkscape.org
%% Accompanies image file 'motivating_example.pdf' (pdf, eps, ps)
\begingroup%
  \makeatletter%
  \providecommand\color[2][]{%
    \errmessage{(Inkscape) Color is used for the text in Inkscape, but the package 'color.sty' is not loaded}%
    \renewcommand\color[2][]{}%
  }%
  \providecommand\transparent[1]{%
    \errmessage{(Inkscape) Transparency is used (non-zero) for the text in Inkscape, but the package 'transparent.sty' is not loaded}%
    \renewcommand\transparent[1]{}%
  }%
  \providecommand\rotatebox[2]{#2}%
  \newcommand*\fsize{\dimexpr\f@size pt\relax}%
  \newcommand*\lineheight[1]{\fontsize{\fsize}{#1\fsize}\selectfont}%
  \ifx\svgwidth\undefined%
    \setlength{\unitlength}{348.52627251bp}%
    \ifx\svgscale\undefined%
      \relax%
    \else%
      \setlength{\unitlength}{\unitlength * \real{\svgscale}}%
    \fi%
  \else%
    \setlength{\unitlength}{\svgwidth}%
  \fi%
  \global\let\svgwidth\undefined%
  \global\let\svgscale\undefined%
  \makeatother%
  \begin{picture}(1,1.01844504)%
    \lineheight{1}%
    \setlength\tabcolsep{0pt}%
    \put(0,0){\includegraphics[width=\unitlength,page=1]{motivating_example.pdf}}%
  \end{picture}%
\endgroup%

%% file: problem.tex
\subsection{Stochastic Games}
%We consider a graph-based game formalism between two players.
A (2.5-player) \emph{stochastic game} (SG) is a tuple $\sg = \langle S,
\iota, \Act, P \rangle$.  The finite set of \emph{states} $S = S_\pOne \cup
S_\pTwo$ is partitioned into a set $S_\pOne$ of (controlled)
$\pOne$-states and a set $S_\pTwo$ of (uncontrolled)
$\pTwo$-states. $\iota \in S_\pOne$ is the \emph{initial state}, $\Act$~is a
finite set of \emph{actions}, and $P\colon S \times \Act \rightarrow
\Distr(S)$ is the \emph{transition function}. For simplicity of
exposition, we assume w.l.o.g. that controlled and uncontrolled
states alternate. Thus, $P$ is defined by two \emph{partial} transition functions:
$P_\pOne\colon S_\pOne \times \Act \rightarrow \Distr(S_\pTwo)$,
$P_\pTwo\colon S_\pTwo \times \Act \rightarrow \Distr(S_\pOne)$. We
identify the available actions\footnote{We use a partial function as we explicitly allow modeling 
unavailable actions, e.g., we can model that a door can only be opened
when close enough to the door.} as $\EnAct(s) \eqdef \{ \act \mid
P(s,\act) \neq \bot \}$. States without available actions, i.e.,
states with $\EnAct(s) = \emptyset$ are called \emph{terminal states}.
The \emph{successor
states} of a state~$s$ and an (enabled) action~$\act$ is the set of
states that are reached from $s$ within one step with a positive
transition probability, i.e., $\Succ(s,\act) \eqdef \{ s' \mid
P(s,\act)(s')>0 \}$, and $\Succ(s) \eqdef\bigcup_{\act \in \EnAct(s)}
\Succ(s,\act)$.

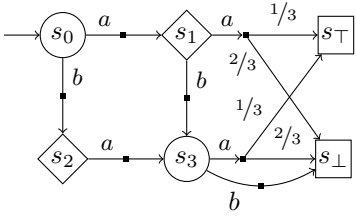
\begin{figure}
\centering
\begin{tikzpicture}
	\node[sstate,initial,initial text=] (s0) {$s_0$};
	\node[astate,right=of s0] (s1) {$s_1$};
	\node[astate,below=of s0] (s2) {$s_2$};
	
	\node[sstate,below=of s1] (s3) {$s_3$};
	
	\node[tstate,right=1.4cm of s1] (target) {$\target$};
	\node[tstate,right=1.4cm of s3] (sink) {$\sink$};
	
	\node[actnode,right=4mm of s1] (a1) {};
	\node[actnode,right=4mm of s3] (a2a) {};

	\draw[->] (s0) -- node[actnode] {} 
	                  node[near start,auto,elab] {$a$} (s1);
	\draw[->] (s0) -- node[actnode] {}
					  node[near start,elab,right] {$b$} (s2);
					  
	\draw[->] (s2) -- node[actnode] {}
					  node[near start,elab,above] {$a$} (s3);				  
					  
	\draw[->] (s1) -- node[actnode] {}
					  node[near start,elab] {$b$}  (s3);
	\draw[->] (s1) -- node[elab] {$a$} (a1);
	\draw[->] (a1) -- node[elab] {$\nicefrac{1}{3}$} (target);
	\draw[->] (a1) -- node[elab,near start,left] {$\nicefrac{2}{3}$} (sink);
	\draw[->] (a2a) -- node[elab,near start,xshift=0.1cm] {$\nicefrac{1}{3}$} (target);
	\draw[->] (a2a) -- node[elab,pos=0.6,above] {$\nicefrac{2}{3}$} (sink);
	\draw[->] (s3) -- node[elab] {$a$} (a2a);
	\draw[->] (s3) edge[bend right=30] node[actnode] {}
										node[near start,below] {$b$} (sink);

\end{tikzpicture}	
\caption{A running example.}
\label{fig:toysg}
\end{figure}
\begin{example}
  We introduce a six-state toy-example~(Fig.~\ref{fig:toysg}) to
  illustrate the definitions. Terminal states are drawn with a rectangle,
  $\pOne$-states with a circle and $\pTwo$-states with a diamond. For
  every state $s$ and action $\act$, we draw transitions in the form
  of edges that connect all successors $s'$, and label them with the
  associated probabilities $P(s,\act)(s')$. For conciseness, we omit
  labelling probability $1$ transitions.
\end{example}

SGs capture a variety of models.  For example, if $|\EnAct(s)| = 1$ for all uncontrolled states, $s \in S_\pTwo$, then $\sg$ is a
\emph{Markov decision process} (MDP).  If $|\EnAct(s)| = 1 $ for all $s \in S$, then $\sg$ is a \emph{Markov chain}. If
$P(s,\act)$ is a Dirac distribution for every $s \in S$ and $\act \in
\Act$, then $\sg$ is called \emph{deterministic} or a \emph{2-player
game}.

\subsection{Paths and Path Properties}
A finite \emph{path}, $\path$, of length $n$ is a sequence $s_0
\xrightarrow{\act_0} s_1 \xrightarrow{\act_1} s_2 \rightarrow \hdots
\rightarrow s_n$ in $\left( S \times \Act \right)^{n} \times S$ where
$P(s_i,\act_i)(s_{i+1}) > 0$ for each $i$.  We denote the length with
$|\path|$, and denote $s_n$, i.e., the last element of $\path$, with
$\last{\path}$. Further, note that $\pOne$ states are even
indexed and $\pTwo$ states are odd indexed as we assume alternation.
A path, $\path' = s'_0 \xrightarrow{\act'_0} \hdots$, is a
\emph{prefix} of~$\path$, if for all $i \leq |\path'|$, $s_i = s'_i$
and for all $i < |\path'|$, $\act_i = \act'_i$.  The set of all finite
paths of length $n$ is denoted $\Paths[n]{\sg}$, and $\Paths{\sg} =
\bigcup_{n \in \NN} \Paths[n]{\sg}$. We omit $\sg$ whenever it is
 clear from the context.
It is helpful to partition paths
based on their last state: $\POnePaths[]{} = \{ \path \in
 \Paths[]{} \mid \last{\path} \in S_\pOne \}$ and $\PTwoPaths[]{} =
\Paths[]{} \setminus \POnePaths[]{}$.

\begin{example}
  In Fig.~\ref{fig:toysg}, there are two paths that end in $s_3$, $s_0 \xrightarrow{a} s_1 \xrightarrow{b} s_3$ and $s_0 \xrightarrow{b} s_2 \xrightarrow{a} s_3$, both of length $2$. Both paths are in $\POnePaths[]{}$, as $s_3 \in S_\pOne$.
\end{example}
%\paragraph{Policies.} 
Whenever some state $s$ is reached, the corresponding player draws an action from $\EnAct(s)$. As standard, we capture this with the notion of a scheduler\footnote{Also known as \emph{strategy} or \emph{policy}.}.
A \emph{scheduler} is a tuple of \emph{player policies} $\sched = \langle \sched_\pOne, \sched_\pTwo \rangle$
with $\sched_{\player} \colon \PlayerPaths[]{} \rightarrow \Distr(\Act)$ such that $\supp(\sched_i(\path)) \subseteq \EnAct(\last{\path})$ for each $\path$, i.e., for every history, the policy sets a distribution over the enabled successor actions.
For a given path, $\path$ and a policy $\sched_{\player}$, we denote by $\sched_{\player}(\act~|~\path)$ the distribution of actions induced by $\sched_{\player}$ given the path $\path$.
%We refer to $\sched_i$, $i \in \{ 1, 2 \}$ as a \emph{Player-i policy}. 
%We denote the $\pOne$-policy $\pOneSched$ and the $\pTwo$-policy $\pTwoSched$ with $\sched_\pOne$ and $\sched_\pTwo$, respectively.
To ease notation, we liberally use the notation $\sched \colon \Paths{} \rightarrow \Distr(\Act)$, where this function is given dependent on which player owns the last state.
 
%
%Applying a policy $\sched$ to an SG $\sg$ yields an \emph{induced Markov chain} $\induced{\sg}{\sched} = \langle S', \iota', P' \rangle$ with state space $S' = \Paths{\sg}$, initial state $\iota' = \iota$, and transition function $P'(\path,\path')$ defined by $P'(\path)(\path \cdot \act s') = \sched(\path)(\act) \cdot P(\last{\path},\act)(s')$ and $P'(\path,\path') = 0$ otherwise. For any upper bound on the length of the paths, the induced MC is finite. 

\begin{example}
  An example for a $\pOne$-policy $\pOneSched$ is given
  by,
  \begin{equation*}
    \pOneSched(\alpha~|~\xi) =
    \begin{cases}
      \nicefrac{1}{2} & \text{if } \alpha \in \{ a,b \}, ~\xi = s_0,\\
      1 & \text{if } \alpha = a,~\xi = s_0\xrightarrow{b}s_2 \xrightarrow{a} s_3,\\
      1 & \text{if } \alpha = b,~\xi = s_0\xrightarrow{a}s_1\xrightarrow{b}s_3.\\
    \end{cases}
  \end{equation*}
\end{example}

%\paragraph{Properties.}
The probability $\Pr(\path \mid \sched)$ of a finite path $\path$ in an SG $\sg$ conditioned on a policy $\sched$ is given by the product of the transition probabilities along a path. 
More precisely, we define the probability $\Pr(\path \mid \sched)$ recursively as:
\begin{equation}
  \begin{split}
    \Pr(s \mid \sched) &\eqdef 1\\
    \Pr(\path~|~\sigma) &\eqdef \Pr(\path'\mid\sched)\cdot \sched(\act\mid\path') \cdot P(\last{\path'},\act)(s')
  \end{split}
\end{equation}
where $\path =  \path' \xrightarrow{\act} s'$.
The probability of a prefix-free set $X \subseteq \Paths{}$  of paths is the sum over the individual path probabilities, $\Pr(X \mid \sched) = \sum_{\path \in X} \Pr(\path \mid \sched )$.

Next, we develop machinery to distinguish between desirable and
undesirable paths. We focus on finite path properties,
referred to as specifications or constraints, that are decidable
within some fixed $\tau \in \Nat$ time steps, e.g., ``Recharge before
t=20.'' Technically, we represent these path properties as prefix free
sets of finite paths, $\soft$, reflecting some formal
property\footnotemark. An example are all paths that end in a particular terminal state~$\target$ within $\tau$ steps.
%In all definitions, we omit the superscript $\sg$ whenever it is clear from the context.

\footnotetext{{Such paths may e.g. be defined using temporal properties such as linear temporal logic over finite traces (LTLf)~\cite{DBLP:conf/ijcai/GiacomoV13}.}}

\subsection{Control Improvisation}
In control improvisation, we aim to find an $\pOne$-policy,
$\pOneSched$, that satisfies a combination of hard- and soft
constraints, and additionally generates surprising behavior, where we
measure the expected surprise by the causal
entropy~\cite{DirectedInfoTheoery} over the paths.

We first define causal entropy on arbitrary sequences of random variables.
Let $\rv{X}_{1:i} \eqdef \rv{X}_1, \hdots, \rv{X}_i$ and $\rv{Y}_{1:i} \eqdef
\rv{Y}_1,\hdots,\rv{Y}_i$ denote two sequences of random variables. The
probability of $ \rv{X}_{1:i}$ causally conditioned on $\rv{Y}_{1:i}$ is:
\begin{equation}
  \causalprob{\rv{X}_{1:i}}{\rv{Y}_{1:i}} \eqdef \prod_{j=1}^i \Pr(\rv{X}_j \mid \rv{X}_{1:j-1}\rv{Y}_{1:j}).
\end{equation}
The causal entropy of $\rv{X}_{1:i}$ given $\rv{Y}_{1:i}$ is then defined as,
\begin{equation}
  H(\rv{X}_{1:i}\mid\mid \rv{Y}_{1:i}) \eqdef \expOver{\rv{X}_{1:i},\rv{Y}_{1:i}}{-\log(\causalprob{\rv{X}_{1:i}}{\rv{Y}_{1:i}})}
\end{equation}
Using the chain rule, one can relate causal entropy to (non-causal) entropy, $H(\rv{X} | \rv{Y}) \eqdef \expOver{\rv{X}}{-\log(\Pr(\rv{X}~|~\rv{Y}))}$ via:
\begin{equation}
  H(\rv{X}_{1:i}\mid\mid \rv{Y}_{1:i}) = \sum_{t=1}^i H(\rv{X}_t \mid \rv{Y}_{1:t}, X_{1:t-1})
\end{equation}
This relation shows that:
(1)~Causal entropy is always lower bounded by non-causal entropy (and
thus non-negative). 
(2)~Causal entropy can be computed ``backward in time''.
(3)~Causal and non-causal conditioning can be mixed,

\vspace{-1.1em}
\begin{equation}
  H(\rv{X}_{1:i}\mid\mid \rv{Y}_{1:i} \mid Z) \eqdef \sum_{t=1}^i H(\rv{X}_t \mid \rv{Y}_{1:t}, X_{1:t-1}, Z).
\end{equation}
    
Intuitively, and contrary to non-causal entropy, causal entropy does \emph{not}
condition on variables that have not been revealed, e.g., on events in the future. 
This makes causal entropy particularly well suited for measuring
predictability in \emph{sequential} decision making problems, as the 
agents cannot observe the future~\cite{mceThesis}.

We now define causal entropy in stochastic games.
Recall that a path alternates states and actions.  The
next state after observing a sequence of state-action pairs is a
random variable. Formally, given $\sg$ and a scheduler 
$\sched$, let us denote by $\rv{A}^{\pOne}_{1:i}$ and
$\rv{S}_{1:i}$ random variable sequences for
$\pOne$-player actions and states respectively. The causal entropy of
controllable actions in $\tau$-length paths under $\sched$ is then,
\begin{equation}
  H_\tau(\sigma) \eqdef H( \rv{A}^{\pOne}_{1:\tau'} \mid\mid \rv{S}_{1:\tau} ),
\end{equation}
where $\tau' = \lceil \frac{\tau}{2}\rceil$ is the number of $\pOne$-actions due to alternation.

%, which we liberally use the set $\Paths[\tau]{\sg \mid \sched}$ to indicate these sequences (and the distribution over them) .  We want to express the causal entropy of a sequence of $\pOne$-action choices given the paths. 
% For that, let us define $\Act_\pOne(\path)$ as the sequence $\act_{j_1} \act_{j_2} \hdots \act_{j_n}$ from $\path = s_0\act_0\hdots s_n$ where $j_i$ are the indices such that $s_{j_i} \in S_\pOne$, i.e., 
% $\Act_\pOne(\path)$ denotes the sequence of action choices of $\pOne$. We can lift the operation to sets: $\Act_\pOne(X) = \{ \Act_\pOne(\path) \mid \path \in X \}$, which we also use to denote the associated random variables.
%\[ H^\sg_\tau(\sigma) \colonequals H( \Act_\pOne(\Paths[\tau]{\sg \mid \sched})   \mid\mid \Paths[\tau]{\sg \mid \sched} ).  \]

\begin{example}
  Consider the uniform $\pOne$ policy on Fig.~\ref{fig:toysg}. If $\sched_{\pTwo}(a\mid \xi) = 1$. $H_\tau(\sigma) = \log(2) + \nicefrac{1}{2}(\log(2))$. Note, only $\pOne$ can \emph{add} entropy, while $\pTwo$ and stochastic transitions yield convex combinations via expectation.
\end{example}

\noindent
We now formalize the problem statement. 
\begin{mdframed}[backgroundcolor=blue!5,nobreak=true]
\textbf{The Entropic Control Improvisation (ERCI) Problem}:
Given a SG $\sg$, $\tau$-bounded path properties $\hard$ and $\soft$, and thresholds $\scthreshold \in [0,1]$ and $\randomness \in [0,\infty)$, find a $\pOne$-policy $\pOneSched$ (or report that none exists) such that for every $\pTwo$-policy $\pTwoSched$,
\begin{compactenum}
	\item (\emph{hard constraint}) $\Pr(\hard \mid \sched) \geq 1$
	\item (\emph{soft constraint)} $\Pr(\soft \mid \sched) \geq \scthreshold$
\item (\emph{randomness constraint}) $H_\tau(\sigma) \geq \randomness$
\end{compactenum}
where  $\sched = \langle \pOneSched, \pTwoSched \rangle$.
\end{mdframed}
We say that an instance of the ERCI problem is realizable, if an appropriate $\pOneSched$ exists and call such $\pOneSched$ an \emph{improviser}. The problem is unrealizable otherwise.

%%% Local Variables:
%%% mode: latex
%%% TeX-master: "main"
%%% End:

%% file: aspects.tex
\section{ERCI as multi-objective optimization}\label{sec:convex}
We investigate the ERCI problem statement. Based on a sequence of observations, we reduce the ERCI problem to the Core ERCI problem which significantly eases the description (and implementation) of the algorithm afterwards.

\subsection{Preprocessing}
To ease the technical exposition, without loss of generality, we make
the following assumptions:
We assume the graph
structure underlying the SG is finite and acyclic -- and thus all paths
are finite length. When considering $\tau$-bounded path properties (monitorable by finite automata),
this assumption is naturally realized by a $\tau$-step unrolling of a
monitor augmented SG \footnotemark, i.e., augmenting the state space with a counter from $0$ to $\tau$ and the current property monitor state. 

\footnotetext{
One may then represent this unrolled graph as a binary
decision diagram, resulting in a (typically) concise graph that grows
proportional to the horizon and minimal state space augmentation
required~\cite{DBLP:conf/cav/Vazquez-Chanlatte20}.}

Next, in order to ensure the hard constraint, $\hard$, we
calculate all states from which the $\pTwo$-player can enforce
violating the hard constraint. Such states are identifiable using a single
topologically ordered pass over $\sg$ from the terminal states to the initial state.  We remove
such states along with their in- and outgoing transitions. Any
$\pOne$-policy now satisfies the hard constraint. 
The remaining terminal states are all merged into two states $\target$ and
$\sink$, based on membership in $\soft$, i.e.,
\begin{equation}
  \begin{split}
    \last{\path} = \target ~\implies~ \path \in \soft\\
    \last{\path} = \sink ~\implies~ \path \notin \soft
  \end{split}.
\end{equation}

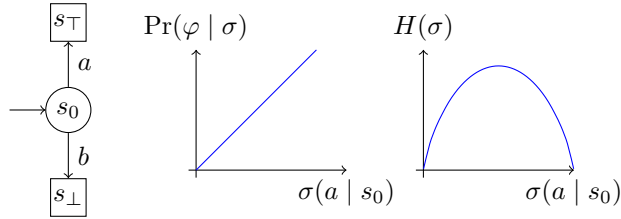
\begin{figure}
\centering
\begin{subfigure}{0.24\columnwidth}
\centering
\begin{tikzpicture}	
	\node[sstate,initial, initial text=] (si) {$s_0$};
	\node[tstate,above=0.6cm of si] (s0) {$\target$};
	\node[tstate,below=0.6cm of si] (s1) {$\sink$};
	\draw[->] (si) -- node[right] {$a$} (s0);
	\draw[->] (si) -- node[right] {$b$} (s1);
	
\end{tikzpicture}
\caption{Minimal MDP}
\label{fig:minimal:mdp}
\end{subfigure}
\begin{subfigure}{0.36\columnwidth}
\centering
\begin{tikzpicture}[scale=2]
 \draw[->] (-0.05, 0) -- (1, 0) node[below]{$\sched(a \mid s_0 )$};
  	\draw[->] (0, -0.05) -- (0, 0.8) node[above] {$\Pr( \varphi \mid \sched)$};
  	%\draw[-,dashed] (0.4,0) -- (0.4,0.5184);
  	%\draw[-,dashed] (0.0,0.5184) -- (0.4,0.5184);
  \draw[ domain=0:0.8, smooth, variable=\x, blue] plot ({\x}, {\x});
\end{tikzpicture}
\caption{Probability to reach $\target$}
\label{fig:minimal:soft}
\end{subfigure}
\begin{subfigure}{0.36\columnwidth}
\centering
\begin{tikzpicture}[scale=2]	
 \draw[->] (-0.05, 0) -- (1, 0) node[below]{$\sched(a \mid s_0 )$};
  	\draw[->] (0, -0.05) -- (0, 0.8) node[above] {$H(\sched)$};
  	%\draw[-,dashed] (0.4,0) -- (0.4,0.5184);
  	%\draw[-,dashed] (0.0,0.5184) -- (0.4,0.5184);
  \draw[ domain=0.001:0.999, smooth, variable=\x, blue] plot ({\x}, {-\x * ln(\x) - (1 - \x)*  ln(1-\x)});
\end{tikzpicture}
\caption{Causal Entropy}
\label{fig:minimal:entropy}
\end{subfigure}

\caption{Minimal ERCI problem with $\varphi = ( \last{\xi} = \target )$}
\end{figure}

\begin{example}
	In Fig.~\ref{fig:minimal:mdp} we show a (deterministic) MDP and we plot for all schedulers the induced probability to reach $\target$ and the induced causal entropy, in Fig.~\ref{fig:minimal:soft} and \ref{fig:minimal:entropy}, respectively. 
	We see that taking action $a$ with increasing probability yields a larger probability to reach $\target$, whereas taking action $a$ and~$b$ uniformly at random is optimal for the entropy. 
\end{example}

\input{geometric}
\subsection{Geometric Perspective}
There is a natural trade-off
between probability of generating paths in $\varphi$ (from here
onwards: \emph{the performance}) and causal entropy induced by a
policy (\emph{the randomization}).  In particular, with all other ingredients fixed, 
we are interested in understanding the combinations of $\scthreshold$
and $\randomness$ that yield a solvable instance of the (core) ERCI problem. To this
end, we cast ERCI as an instance of a multi-objective optimization problem, and
study its Pareto front. Some ideas are inspired by variants of multi-objective analysis of MDPs with multiple soft constraints, e.g.~\cite{DBLP:conf/stacs/ChatterjeeMH06,DBLP:conf/tacas/EtessamiKVY07,DBLP:conf/atva/ForejtKP12}.

It is convenient to consider this front geometrically.
To begin, given a fixed ERCI instance, a scheduler $\sched$
\emph{induces} a point $x_\sched$:
\begin{equation}
  x_\sched \eqdef \Big\langle \Pr(X_\varphi \mid \sched), H(\sched) \Big\rangle \in [0,1] \times [0,\infty).  
\end{equation}
To ease notation, for $x_\sched = \langle \scp,\rndp \rangle$ we use
$\scp_\sched \eqdef \scp$ and $\rndp_\sched \eqdef \rndp$. Next, we
partially order these points via the standard product ordering:
\begin{equation}
  \langle \scp,\rndp \rangle \preceq \langle \scp',\rndp' \rangle \quad\text{ iff }\quad \scp \leq \scp' \wedge \rndp \leq \rndp'.
\end{equation}

We say that $\pOneSched$ \emph{guarantees} a point $x_\pOne \eqdef
\langle \scp, \rndp \rangle$, if for every policy $\pTwoSched$, using
$\sched = \langle \pOneSched, \pTwoSched \rangle$, we have
$\scp_\sched \geq \scp$ and $\rndp_\sched \geq \rndp$. Thus, a point
is guaranteed if no matter what policy $\pTwo$ uses, $x_\sched$ will
induce a point no worse w.r.t.\ to either randomization or performance
than $x_\pOne$. 
We define
\emph{the set of guaranteed points} for a scheduler $\pOneSched$:
\begin{equation}\label{eq:guaranteed}
  \solutions[\pOneSched] \eqdef \{ \langle \scp, \rndp \rangle \mid  \pOneSched \text{ guarantees } \langle \scp, \rndp \rangle \}.
\end{equation}
We observe that guaranteed points are
downward closed, i.e., if $\pOneSched$ guarantees $x$ and $x' \preceq x$,
then $\pOneSched$ guarantees $x'$.
\begin{example}
Consider Fig.~\ref{fig:geom:guarantee}. We fix $\pOneSched$ and in the blue hatched area draw all points induced by $\sched = \langle \pOneSched, \pTwoSched \rangle$ when varying $\pTwoSched$. We take the minimal randomness $\rndp$ and the minimal performance $\scp$. The points in the downward closure  of $\langle \scp, \rndp \rangle$ (green circle) are the guaranteed points for $\pOneSched$ in the green solid area.	
We notice the gap between both areas: While the performance and randomization may be better than the optimum that $\pOne$ can guarantee, it cannot guarantee a higher randomization \emph{and} performance simultaneously, as  the $\pTwo$-player would have a counter-policy violating either the performance \emph{or} the randomization.
\end{example}

Points guaranteed by some $\pOneSched$ are called
\emph{achievable}. Thus, the achievable points are: $ \solutions =
\bigcup_{\pOneSched} \solutions[\pOneSched]$.  Importantly, the ERCI problem is realizable iff $\langle \scthreshold,
\randomness \rangle$ is achievable. 
Thus, to solve ERCI instances, we start by characterizing
$\solutions$. We start by observing the $\solutions$ is convex\footnotemark~(proof in Sec~\ref{sec:proofs}).

\begin{proposition}\label{prop:convex}
  The set of achievable points, $\solutions$, is convex. 
\end{proposition}

\footnotetext{
  That is, $x, x' \in \solutions$ implies for
  every $w \in [0,1]$ that $w \cdot x + (1-w) \cdot x \in \solutions$
}

Next, because $\solutions$ is downward closed, it
suffices to study the ``maximal'' or non-dominated points.  Precisely,
we say that a point $x$ is \emph{dominated} by $x'$ if $x \prec
x'$, i.e., if $x \preceq x' \wedge x \neq x'$.
The Pareto front $\pareto{\solutions}$ of $\solutions$ is then the set of non-dominated achievable  points,
\begin{equation}
  \pareto{\solutions} \eqdef \{ x \in \solutions \mid \forall x' \in \solutions, x \not\prec x'  \}.  
\end{equation}
\noindent
\begin{mdframed}
Importantly, it holds that the ERCI problem is satisfiable iff there exists a  $x \in \pareto{\solutions}$ such that $\langle \scthreshold, \randomness \rangle \preceq x$.    
\end{mdframed}
\begin{example}
	The set $\solutions$ illustrated in Fig.~\ref{fig:geom:solution} is obtained by taking the union of guaranteed points, and can be characterized by the set of points on the Pareto front: This is the curved border between the green and white area, in particular the three green dots are on the Pareto front. Any ERCI instance with $\langle \scthreshold, \randomness \rangle$ in the green area is realizable.
\end{example}

\noindent
Approximating the Pareto front gives a natural approximation
scheme for ERCI instances: For any subset $\pareto{} \subseteq
\pareto{\solutions}$,
\begin{enumerate}
\item If there exists an $x \in \pareto{}$ such that
$\langle \scthreshold, \randomness \rangle \preceq x$, then the ERCI
problem must be realizable and $x$ is a witness to realizability.
\item If there exists an $x \in \pareto{}$
such that $x \prec \langle \scthreshold, \randomness \rangle$, then the
ERCI problem is not realizable and $x$ is a witness to unrealizability.
\end{enumerate}
Due to convexity, we may speed up the search for realizability: If there exist $x_1, x_2 \in \pareto{}$ such that $\langle \scthreshold, \randomness \rangle \prec \big(w \cdot x_1 + (1-w) \cdot x_2\big)$, we call $x_1,x_2$ a witness-pair.
\begin{remark}
Given a witness(pair) to realizability, it is easy to extract the corresponding improviser. Let $x_1,x_2$ be a witness-pair to realizability,  induced by $\sched_{\rat_1}$ and $\sched_{\rat_2}$ such that $\langle\scthreshold, \randomness\rangle\preceq w\cdot x_1 + (1-w)\cdot x_2$, then the policy described by  	
\begin{equation}\label{eq:4}
  \sigma^*_\pOne(\act \mid s) \eqdef w\cdot \sched_{\rat_1}(\act \mid s) + (1-w) \cdot \sched_{\rat_2}(\act  \mid s)
\end{equation}
is an improviser solving the ERCI problem.
\end{remark}

\begin{example}
\label{ex:approximation}
	Consider Fig.~\ref{fig:geom:iterative}. We have found three points on the Pareto front, and already have a good impression of the trade-off between randomization and performance. In particular, the green area is definitively a subset of $\solutions$: It exploits the downward closure and the convexity of $\solutions$. The red (dotted) part contain the points on the Pareto front in their downward closure, thus they cannot be part of the Pareto front themselves.
	Furthermore, the topmost point on the Pareto front was obtained by maximizing performance (and optimizing randomization only as a secondary objective). Thus, by construction, the bricked area at the top is not realizable. Analogously, the bricked area at the right reflects non-achievable randomization. 
\end{example}

\begin{remark}\label{rem:scalarwitnesses}
We notice that the multi-objective optimization perspective allows us to extend the set of witnesses for unrealizability. In particular, every point of the Pareto-curve can be described as optimizing some scalarization of the objectives. Geometrically, it optimizes along a particular direction. Whenever we know that a Pareto-optimal point $x = \langle \scp, \rndp \rangle$ optimizes a weighted objective with weights $w = \langle w_1, w_2 \rangle$, then $x$ and $w$ \emph{together} are a witness for unrealizability for $\langle \scthreshold, \randomness \rangle$ whenever $w_1 \cdot \scp + w_2 \cdot \rndp < w_1 \cdot \scthreshold + w_2 \cdot \randomness$. 
\end{remark}

Thus a key algorithmic question in ERCI is how to efficiently explore
 the Pareto front $\pareto{\solutions}$. 
 
 \subsection{Regret-Based ERCI}
 To algorithmically explore the Pareto-curve, we re-parameterize the ERCI problem. 
 
 First, we find  the two special points induced by (1) optimizing performance and only then randomization (the topmost green point in the figures) and (2) optimizing randomization and only then performance (the rightmost green point). 
As we have seen, these restrict the domain in which we can actually trade performance for randomness. 
We define 
$\rndopt \eqdef \max \{ \rndp \mid \exists \scp \text{ s.t. } \langle \scp, \rndp \rangle \in \solutions  \} $, i.e., the largest randomness that can be guaranteed by any $\pOne$-policy. 
Likewise, we define 
$\scopt \eqdef \max \{ \scp \mid \exists \rndp \text{ s.t. } \langle \scp, \rndp \rangle \in \solutions  \} $, i.e., the largest performance that can be guaranteed by any $\pOne$-policy. 
Then, we define 
$\scmin \eqdef \max \{ \scp \mid \langle \scp, \rndopt \rangle  \in \solutions \}$, the best performance that $\pOne$ can guarantee while guaranteeing optimal randomness. 
Likewise, we define  the analogous $\rndmin \eqdef \max \{ \rndp \mid \langle \scopt, \rndp \rangle  \in \solutions \}$.
We thus obtain two points on the Pareto front: $\langle \scmin, \rndopt \rangle$ and $\langle \scopt, \rndmin \rangle$, and intuitively, we can trade between these two points following the Pareto front.

%\begin{remark}[Regret Based ERCI]
Now, rather that fixing $\scthreshold$ and $\randomness$ a
  priori, we seek to guarantee some percentage of the independently achievable soft
  constraint and causal entropy measure.
 We
  re-parameterize ERCI as follows:
  \begin{equation}
    \scthreshold_\epsilon \eqdef \epsilon \cdot (\scopt - \scmin) + \scmin
    \hspace{1em}
    \randomness_\delta \eqdef  \delta \cdot (\rndopt - \rndmin) + \rndmin
  \end{equation}
  where $\epsilon, \delta \in [0, 1]$. We call this version of ERCI \emph{regret-based}. We remark that the reparameterization is not only beneficial from a usability point-of-view, but it also eases our exposition.  Geometrically, after computing $\scopt$ and $\rndopt$, we know that the left triangle in Fig.~\ref{fig:geom:regret} is definitively realizable, and the regret-based ERCI asks whether the white circle is also realizable (where the point of the white point is given by $\epsilon$ and $\delta$. %As we shall later see, these maximum quantities are
 % directly computed in our proposed algorithm. 
%\end{remark}
\noindent 
Together, we obtain the following (core) ERCI problem.
\begin{mdframed}[backgroundcolor=blue!5,nobreak=true]
\textbf{The Core ERCI Problem}:
Given an finite acyclic SG~$\sg$, with terminal states, $\target$ and $\sink$, and thresholds $\epsilon, \delta \in [0,1]$,  find a $\pOne$-policy $\pOneSched$ s.t.~for every $\pTwo$-policy~$\pTwoSched$:
\begin{enumerate}
\item (\emph{soft constraint)}
  $\Pr(\last{\xi} = \target \mid \sched) \geq \scthreshold_\epsilon$
\item (\emph{randomness constraint}) $H(\sigma) \geq \randomness_\delta$
\end{enumerate}
where  $\sched = \langle \pOneSched, \pTwoSched \rangle$.
\end{mdframed}

Finally, it is helpful to think about the Pareto front as a function of randomization in this reparameterization.  We define a characteristic function which given a target
performance ratio, $\epsilon$, yields the optimal randomness ratio,
$\delta$:
\begin{equation}
  \begin{split}
    & \solfuncp\colon [0,1] \rightarrow [0, 1]    \\
    & \solfuncp(\delta) = \max_\epsilon \{ \randomness_\delta \mid \langle
    \scthreshold_\epsilon, \randomness_\delta \rangle \in \solutions \} 
  \end{split}
\end{equation}
\begin{proposition}\label{prop:monotone}
  $\solfuncp$ is continuous and (strictly) decreasing.
\end{proposition}
 We shall temporarily postpone the proof of
\propref{monotone}. For now, one case observe that
(non-strict) monotone decreasing follows directly from convexity and
using the adequate domains.
Finally, the set  $\solutions$ is (in general) \emph{not} a finite polytope -- the MDP in Fig.~\ref{fig:minimal:mdp} serves as an example. Nevertheless,  $\solutions$ can be well approximated with finitely many vertices, see Ex.~\ref{ex:approximation}.

With these facts, we are now well-equipped to develop the algorithms in Sec.~\ref{sec:mdps} for MDPs and Sec.~\ref{sec:sgs} for SGs.

%%% Local Variables:
%%% mode: latex
%%% TeX-master: "main"
%%% End:

%% file: geometric.tex
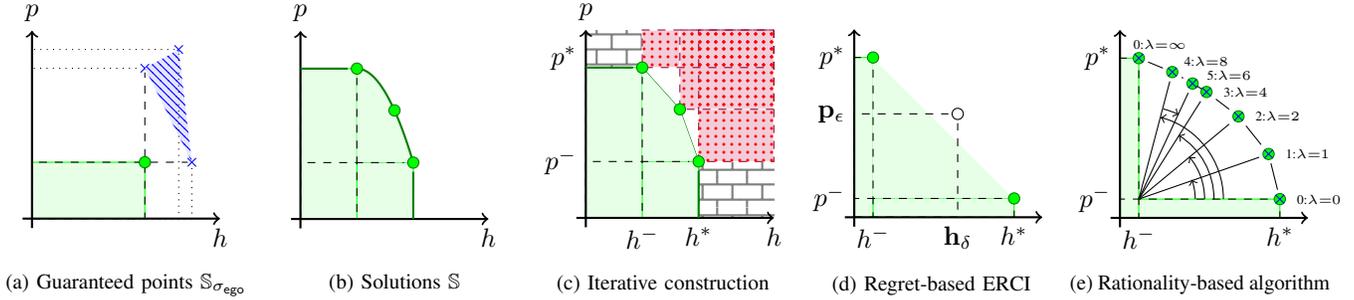
\begin{figure*}
\begin{subfigure}{0.19\textwidth}
\centering
\begin{tikzpicture}[scale=2.5]	
     	
  	\draw (0.85, 0.3) node[cross=2pt,color=blue] (c1) {};
  	\draw (0.78, 0.9) node[cross=2pt,color=blue] (c2) {};
  	\draw (0.6, 0.8) node[cross=2pt,color=blue] (c3) {};
  	
  	\node at (0.6,0.3) (x1) {};
  		\draw[fill=green!10!white,draw=green] (0,0) rectangle (x1);
  		 \draw[thick, ->] (-0.05, 0) -- (1, 0) node[below]{$\rndp$};
  	\draw[thick, ->] (0, -0.05) -- (0, 1) node[above] {$\scp$};

\fill[fill=blue!10] (c1.center)--(c2.center)--(c3.center);
	
	  \fill[fill=blue!20,pattern=north west lines,pattern color=blue] (c1.center)--(c2.center)--(c3.center);

  	\draw[dashed] (0,0) |- (c1);
  	\draw[dotted] (0,0) |- (c2);
  	\draw[dotted] (0,0) |- (c3);
  	\draw[dotted] (0,0) -| (c1);
  	\draw[dotted] (0,0) -| (c2);
  	\draw[dashed] (0,0) -| (c3);

  	\draw (x1) node[circle,fill=green, inner sep=1.5pt,draw=green!50!black] {};

\end{tikzpicture}
\caption{Guaranteed points $\solutions_\pOneSched$}
\label{fig:geom:guarantee}
\end{subfigure}
\begin{subfigure}{0.19\textwidth}
\centering
\begin{tikzpicture}[scale=2.5]	
     	
  	\node at (0.6,0.3) (x1) {};
  		\node at (0.3,0.8) (x2) {};
  	
  		\node at (0.5,0.57778) (x3) {};
  	
  		\draw[fill=green!10!white,draw=green] (0,0) rectangle (x1);
  			\draw[fill=green!10!white,draw=green] (0,0) rectangle (x2);

  		 \draw[thick, ->] (-0.05, 0) -- (1, 0) node[below]{$\rndp$};
  	\draw[thick, ->] (0, -0.05) -- (0, 1) node[above] {$\scp$};

    \fill[ domain=0.3:0.6, smooth, variable=\x, green!10!white,thick] plot ({\x}, {-5.555*(\x-0.3)*(\x-0.3) + 0.8}) -- (0.3,0.3);
    
  		\draw[name path=f, domain=0.3:0.6, smooth, variable=\x, green!50!black,thick] plot ({\x}, {-5.555*(\x-0.3)*(\x-0.3) + 0.8});

  	\draw[dashed] (0,0) -| (x1);
  	\draw[dashed] (0,0) -| (x2);
  	 \draw[dashed] (0,0) |- (x1);
  	\draw[dashed] (0,0) |- (x2);
  	
  		\draw[-,thick,color=green!50!black] (0,0.8) -- (x2);
  	\draw[-,thick,color=green!50!black] (0.6,0.0) -- (x1);
  	 
  	\draw (x1) node[circle,fill=green, inner sep=1.5pt,draw=green!50!black] {};
  \draw (x2) node[circle,fill=green, inner sep=1.5pt,draw=green!50!black] {};
  \draw (x3) node[circle,fill=green, inner sep=1.5pt,draw=green!50!black] {};

\end{tikzpicture}
\caption{Solutions $\solutions$}
\label{fig:geom:solution}
\end{subfigure}
\begin{subfigure}{0.19\textwidth}
\centering
\begin{tikzpicture}[scale=2.5]	
     	
  	\node at (0.6,0.3) (x1) {};
  		\node at (0.3,0.8) (x2) {};
  	
  		\node at (0.5,0.57778) (x3) {};
  		
  					\draw[draw=none,pattern= bricks, pattern color=black!50] (1,0) rectangle (x1);
  					
  					\draw[draw=none,pattern= bricks, pattern color=black!50] (0,1) rectangle (x2);
  	
  		\draw[fill=green!10!white,draw=green] (0,0) rectangle (x1);
  			\draw[fill=green!10!white,draw=green] (0,0) rectangle (x2);
  			
  				\draw[fill=purple!20!white,draw=red!0!white] (1,1) rectangle (x1);
  			\draw[fill=purple!20!white,draw=red!0!white] (1,1) rectangle (x2);
  					\draw[fill=purple!20!white,draw=red!0!white] (1,1) rectangle (x3);
  						\draw[draw=purple!80!black,dashed] (1,1) rectangle (x1);
  			\draw[draw=purple!80!black,dashed] (1,1) rectangle (x2);
  					\draw[draw= purple!80!black,dashed] (1,1) rectangle (x3);
  					
  					\draw[draw=none,dashed,pattern=dots, pattern color=red] (1,1) rectangle (x1);
  					\draw[draw=none,dashed,pattern=dots, pattern color=red] (1,1) rectangle (x2);
  					\draw[draw=none,dashed,pattern=dots, pattern color=red] (1,1) rectangle (x3);

  		 \draw[thick, ->] (-0.05, 0) -- (1, 0) node[below]{$\rndp$};
  	\draw[thick, ->] (0, -0.05) -- (0, 1) node[above] {$\scp$};

  	 \fill[draw=green!50!black] (x2.center) -- (x3.center) -- (x1.center);
    \fill[color=green!10!white] (x2.center) -- (x3.center) -- (x1.center) --  (0.3,0.3);
    
%  		\draw[name path=f, domain=0.3:0.6, smooth, variable=\x, green!50!black,thick] plot ({\x}, {-5.555*(\x-0.3)*(\x-0.3) + 0.8});
%    
  	
  	\draw[dashed] (0,0) -| (x1);
  	\draw[dashed] (0,0) -| (x2);
  	 \draw[dashed] (0,0) |- (x1);
  	\draw[dashed] (0,0) |- (x2);
  	
  		\draw[-,thick,color=green!50!black] (0,0.8) -- (x2);
  	\draw[-,thick,color=green!50!black] (0.6,0.0) -- (x1);
  	 
  	\draw (x1) node[circle,fill=green, inner sep=1.5pt,draw=green!50!black] {};
  \draw (x2) node[circle,fill=green, inner sep=1.5pt,draw=green!50!black] {};
  \draw (x3) node[circle,fill=green, inner sep=1.5pt,draw=green!50!black] {};
  \node[anchor=north] at (0.6,0) {$\rndopt$};
  		
  		\node[anchor=north] at (0.3,0) {$\rndmin$};
  		
  		\node[anchor=east] at (0,0.85) {$\scopt$};
  		
  		\node[anchor=east] at (0,0.3) {$\scmin$};

\end{tikzpicture}
\caption{Iterative construction}
\label{fig:geom:iterative}
\end{subfigure}
\begin{subfigure}{0.19\textwidth}
\centering
\begin{tikzpicture}[scale=2.5]	
     	
  	\node at (0.85,0.1) (x1) {};
  		\node at (0.1,0.85) (x2) {};
  		
  		\node at (0.55,0.55) (x3) {};
  		
  		\node[anchor=north] at (0.85,0) {$\rndopt$};
  		\node[anchor=north] at (0.55,0) {$\randomness_\delta$};
  		
  		\node[anchor=north] at (0.1,0) {$\rndmin$};
  		
  		\node[anchor=east] at (0,0.85) {$\scopt$};
  		\node[anchor=east] at (0,0.55) {$\scthreshold_\epsilon$};
  		
  		\node[anchor=east] at (0,0.1) {$\scmin$};
  	
  		\draw[fill=green!10!white,draw=green] (0,0) rectangle (x1);
  			\draw[fill=green!10!white,draw=green] (0,0) rectangle (x2);
  			
  			 \fill[draw=white,fill=green!10!white] (x1.center) -- (x2.center) -- (0.1, 0.1);

  		 \draw[thick, ->] (-0.05, 0) -- (1, 0) node[below]{\phantom{$\rndp$}};
  	\draw[thick, ->] (0, -0.05) -- (0, 1) node[above] {\phantom{$\scp$}};
  	
  	\draw[dashed] (0,0) -| (x1);
  	\draw[dashed] (0,0) -| (x2);
  	 \draw[dashed] (0,0) |- (x1);
  	\draw[dashed] (0,0) |- (x2);
  	
  	 \draw[dashed] (0,0) |- (x3);
  	\draw[dashed] (0,0) -| (x3);
  	 
  	\draw (x1) node[circle,fill=green, inner sep=1.5pt,draw=green!50!black] {};
  	
  	\draw (x2) node[circle,fill=green, inner sep=1.5pt,draw=green!50!black] {};

  	\draw (x3) node[circle,draw,fill=white, inner sep=1.5pt,draw=black] {};
  	
\end{tikzpicture}
\caption{Regret-based ERCI}
\label{fig:geom:regret}
\end{subfigure}
\begin{subfigure}{0.19\textwidth}
\centering
\begin{tikzpicture}[scale=2.5]	
     	
  	\node at (0.85,0.1) (x1) {};
  		\node at (0.1,0.85) (x2) {};

  		\node[anchor=north] at (0.85,0) {$\rndopt$};
  		
  		\node[anchor=north] at (0.1,0) {$\rndmin$};
  		
  		\node[anchor=east] at (0,0.85) {$\scopt$};
  		
  		\node[anchor=east] at (0,0.1) {$\scmin$};
  	
  		\draw[fill=green!10!white,draw=green] (0,0) rectangle (x1);
  			\draw[fill=green!10!white,draw=green] (0,0) rectangle (x2);
  			
  		 \draw[thick, ->] (-0.05, 0) -- (1, 0) node[below]{\phantom{$\rndp$}};
  	\draw[thick, ->] (0, -0.05) -- (0, 1) node[above] {\phantom{$\scp$}};
  	
  	\draw[dashed] (0,0) -| (x1);
  	\draw[dashed] (0,0) -| (x2);
  	 \draw[dashed] (0,0) |- (x1);
  	\draw[dashed] (0,0) |- (x2);

  	\draw (x1) node[circle,fill=green, inner sep=1.5pt,draw=green!50!black] {};
  	
  	\draw (x2) node[circle,fill=green, inner sep=1.5pt,draw=green!50!black] {};
  	
  	\draw (x1) node[cross=2pt,color=blue] (c3) {};
  	
  	\draw (x2) node[cross=2pt,color=blue] (c3) {};
  
  \draw[black,->] (0.4,0.1) arc (0:20:0.3cm);
  
  		\node at (0.79,0.34) (x3) {};
  			\draw (x3) node[circle,fill=green, inner sep=1.5pt,draw=green!50!black] {};
  			
  	\draw (x3) node[cross=2pt,color=blue] (c3) {};
  	\draw [-] (0.1,0.1) -- (x3);
   \draw[black,->] (0.45,0.1) arc (0:40:0.35cm);
   
   \node at (0.63,0.54) (x4) {};
  			\draw (x4) node[circle,fill=green, inner sep=1.5pt,draw=green!50!black] {};
  			
  	\draw (x4) node[cross=2pt,color=blue] (c3) {};
  	\draw [-] (0.1,0.1) -- (x4);
   
   \draw[black,->] (0.5,0.1) arc (0:58:0.4cm);
    \node at (0.46,0.67) (x5) {};
  			\draw (x5) node[circle,fill=green, inner sep=1.5pt,draw=green!50!black] {};
  			
  	\draw (x5) node[cross=2pt,color=blue] (c3) {};
  	\draw [-] (0.1,0.1) -- (x5);
   \draw[black,->] (0.55,0.1) arc (0:75:0.45cm);
   \node at (0.278,0.776) (x6) {};
  			\draw (x6) node[circle,fill=green, inner sep=1.5pt,draw=green!50!black] {};
  			
  	\draw (x6) node[cross=2pt,color=blue] (c3) {};
  	\draw [-] (0.1,0.1) -- (x6);
  	
   \draw[black,->] (0.23,0.58) arc (75:65:0.5cm);
   \node at (0.386,0.715) (x7) {};
  			\draw (x7) node[circle,fill=green, inner sep=1.5pt,draw=green!50!black] {};
  			
  	\draw (x7) node[cross=2pt,color=blue] (c3) {};
  	\draw [-] (0.1,0.1) -- (x7);

   \draw[-] (x1) -- (x3) -- (x4) -- (x5) -- (x7) -- (x6) -- (x2);
  
   \node[anchor=west,xshift=0.3em] at (x1) {\tiny{0:$\lambda{=}0$}};
    \node[anchor=south,xshift=0.8em] at (x2) {\tiny{0:$\lambda{=}\infty$}}; 
   \node[anchor=west,xshift=0.3em] at (x3) {\tiny{1:$\lambda{=}1$}};
   \node[anchor=west,xshift=0.3em] at (x4) {\tiny{2:$\lambda{=}2$}};
   \node[anchor=west,xshift=0.3em] at (x5) {\tiny{3:$\lambda{=}4$}};
   \node[anchor=west,yshift=0.3em,xshift=0.2em] at (x7) {\tiny{5:$\lambda{=}6$}};
   \node[anchor=west,yshift=0.4em,xshift=0.1em] at (x6) {\tiny{4:$\lambda{=}8$}};
  	
\end{tikzpicture}
\caption{Rationality-based algorithm}
\label{fig:geom:doubling}
\end{subfigure}

\caption{Geometric interpretation of the ERCI problem for some fixed SG.}
\end{figure*}

%% file: mdp.tex
\section{The Control Improvisation Problem for MDPs}
\label{sec:mdps}

We present an algorithm for the control improvisation problem for
MDPs, which in the next section, will serve as a subroutine for an
algorithm on SGs. Our goal shall be to instantiate the approximation
scheme from the previous section. In particular, we seek
to find points on the Pareto curve $\pareto{\solutions}$ and
incrementally build up $\pareto{} \subseteq \pareto{\solutions}$.
\subsection{Rationality}
To start, recall that an MDP is a stochastic game with no action
choices for the environment, i.e., the environment is purely
stochastic and the only degree of freedom is $\pOne$'s policy.  The
key idea for finding points on the Pareto-curve is to rephrase the
trade-off between randomization and performance as a degree in
rationality $\rat$ of the policy.  Formally, the rationality
corresponds to the following scalarization of our multi-objective
problem~\cite{DBLP:journals/corr/abs-1805-00909},
\begin{equation}
  \label{eq:scalarization}
  J_\rat(\sched) \eqdef \Big\langle 1, \rat\Big\rangle \cdot \Big\langle\rndp_\sched, \scp_\sched\Big\rangle.
\end{equation}
In context of MDPs, the \textbf{unique} ($\pOne$-)policy that
optimizes~\eqref{eq:scalarization} is given by a smooth variant of the
Bellman equations~\cite{mceThesis, DBLP:conf/cav/Vazquez-Chanlatte20}. Namely, let $\smoothmax{}$ denote
the log-sum-exp operator, i.e.,
$\smoothmax(X) \eqdef \log \left( \sum_{x\in X} e^x \right)$. For each
rationality $\rat \in [0, \infty)$, we define a policy $\sched_\rat$
-- using $s = \last{\path}$ -- as follows:
 \begin{align}
   &\sched_\rat(\act \mid s) \eqdef \exp( Q_\rat(s,\act) - V_\rat(s))  \label{eq:mdp:first}\\
   & V_\rat(s) \eqdef  \begin{cases}
     \lambda  \cdot \indicator{s = \target} & \text{if }s \in \{ \target, \sink \},\\
     \smoothmax_{\act \in \EnAct(s)}{  Q_\rat(s,\act) } & \text{otherwise.}
   \end{cases}\label{eq:mdp:v}\\ 
	& Q_\rat(s, \act) \eqdef \sum_{s'} P(s,\act,s') \cdot V_\rat(s').\label{eq:mdp:last}
 \end{align}
To ease notation, we denote $x_\rat \eqdef x_{\sched_\rat},
\scp_\rat \eqdef \scp_{\sched_\rat}, \rndp_\rat \eqdef
\rndp_{\sched_\rat}$. 
 % As previously alluded at the start of the subsection, the key property
% is that $\sched_\rat$ is the \emph{unique} maximum causal entropy policy
% such that $\Pr(\varphi) = \scp_\rat$~\cite{mceThesis}.
Intuitively, as $\rat \rightarrow 0$, $\sched_\rat$ approaches the
uniform distribution over \emph{all available actions}. Note that this
policy maximizes (causal) entropy, and thus $\rndopt = \rndp_0$.  As
$\lambda \rightarrow \infty$, this variant of the Bellman equations
coincides with the standard Bellman
equations~\cite{DBLP:books/wi/Puterman94}, where $\sched_\rat$ selects
(uniformly) from actions \emph{that maximize performance}.
Furthermore, the monotonicity and smoothness of the above Bellman
equations yields the following proposition.
\begin{proposition}
  $\scp_\rat$ is  continuously (and strictly) increasing in $\rat$ and $\rndp_\rat$
  is smoothly (and strictly) decreasing in $\rat$.
\end{proposition}
\noindent In terms of $\solfuncp$, we can define:
\begin{equation}
  \epsilon_\rat \eqdef \frac{\scp_\rat - \scp_0}{\scp_\infty} + \scp_0
  \hspace{1em}\text{and}\hspace{1em}
\delta_\rat \eqdef \frac{\rndp_\rat - \rndp_\infty}{\rndp_0} +
\rndp_{\infty}. 
\end{equation}

Then, because $\sched_\rat$ maximizes randomness
given a target performance, one derives:
\begin{equation}
  \solfuncp\left(\delta_\rat\right) = \epsilon_\rat.
\end{equation}

What remains is to instantiate the approximation scheme for the Pareto
front by varying the optimization direction
$\langle \rat, 1\rangle$.\footnote{Assuming $\scopt, \rndopt \neq 0$
  (which would otherwise yield trivial $\solutions$ and
  $\pareto{\solutions}$)}  In particular, we construct
$\pareto{} = \{ x_\rat \mid \rat \in \{ \rat_1, \rat_2, \hdots \} \}$
until $\pareto{}$ contains a witness to either realizability or
unrealizability of the ERCI instance. We notice that the scalarization in \eqref{eq:scalarization} means that we may additionally exploit witnesses to unrealizability as outlined in Remark~\ref{rem:scalarwitnesses}. In the remainder of this
section, we improve upon randomly selecting values for $\rat$.

%
%and by varying $\rat$ we can explore the Pareto front. First observe the following easily verified proposition.
%\begin{proposition}
%  $\scp_\rat$ is smoothly and (strictly) monotonically increasing in $\rat$ and $\rndp_\rat$
%  is smoothly (strictly) monotonically decreasing in $\rat$.
%\end{proposition}

\subsection{Targeted Pareto-exploration}
The key ingredient to improve upon arbitrarily selecting $\rat_1, \hdots \rat_i$ is to exploit additional structure of the rationality.  
%The key algorithmic idea is thus to strategically evaluate a sequence
%of rationality coefficients to yield (input, output) pairs for
%$\solfuncp$. Due to convexity, the convex hull this sequence of
%rationality-indexed points (and the origin) gradually refines a
%polygonal approximation of $\solutions$, and thus the Pareto
%Front. This approximation, $\hat{\solutions}$, is refined until
%either:
%\begin{enumerate}
%\item $\langle \scthreshold, \randomness \rangle \in \hat{\solutions}$ proving
%  $\langle \scthreshold, \randomness \rangle \in \solutions$.
%\item A $\rat$ is found such that
%  $x_{\rat} \prec \langle \scthreshold, \randomness \rangle$, proving
%  $\langle \scthreshold, \randomness \rangle \notin \solutions$.
%\end{enumerate}
%
%
%Next, to extract an improviser, observe that because
%$\hat{\solutions}$ is a convex polygon, if $\langle \scthreshold,
%\randomness \rangle \in \hat{S}$, then there must two corners of
%$\hat{\solutions}$ indexed by $\rat_1$ and $\rat_2$, that form a
%triangle with $(0, 0)$ containing $\langle \scthreshold, \randomness
%\rangle$. Thus, as in the convexity proof, there must be a convex
%combination of $q\cdot x_{\rat_1} + \bar{q}\cdot x_{\rat_2}$ that
%dominates $\langle \scthreshold, \randomness \rangle$. Therefore, the
%following policy solves the ERCI instance:

%\mypara{Approximation Sequence} 
%The final algorithmic question for
%MDPs is then: what order should one evaluate rationality coefficients.
We propose a three staged sequence: (i) Compute $x_\rat$ for the end
points $\rat \in \{0, \infty\}$.  (ii) Double $\rat$ (starting at $\lambda=1$) until $h_\rat \leq
\randomness$, yielding $\rat_1\ldots \rat_j$.
(iii) Binary search for $\rat \in [\rat_{j-1}, \rat_{j}]$. We illustrate the idea in Fig.~\ref{fig:geom:doubling}.

The algorithm terminates almost surely, that is: 
the algorithm halts if $\langle \scthreshold, \randomness \rangle$ is
not on $\pareto{\solutions}$ (or if we happen to exactly hit $\langle \scthreshold, \randomness\rangle$ by selecting some rationality $\rat$).
As the Pareto front has
measure 0, we argue that not halting is thus merely a technical concern, as a
small perturbation to the ERCI instance (i.e. a \emph{smoothed
analysis}~\cite{SmoothedAnalysis}) on $\sg$ admits decidability. 
\begin{mdframed}
  Our approximation scheme yields a semi-decision process which halts
  iff either (a) $\langle \scthreshold, \randomness \rangle$ is
  bounded away from $\pareto{\solutions}$ \emph{or} (b)
  $\langle \scthreshold, \randomness \rangle$ is dominated by
  $x_{\rat_i}$.
\end{mdframed}
Next, observe that if we terminate the binary search when the search
region is smaller than $\Delta$, this approximation scheme becomes linear in the MDP size
and logarithmic in the final rationality, $\rat_*$, and the
resolution, $\Delta$, i.e., the run-time is,
\begin{equation}
  \mathcal{O}\Big(~\hspace{-1.4em}\underbrace{|\sg|}_{\text{Evaluate $x_\rat$}}\hspace{-1.4em}~\cdot\overbrace{\log(\rat_*)}^{\text{Doubling Phase}}\cdot\underbrace{\log(\nicefrac{1}{\Delta})}_{\text{Binary Search}}\Big)
\end{equation}
Finally, before generalizing to stochastic games, we observe that in
practice, $\rat = 100$ yields a nearly optimal policy, and thus one
can often assume $\rat_* \leq 100$ in our run-time analysis.

%%% Local Variables:
%%% mode: latex
%%% TeX-master: "main"
%%% End:

%% file: sg.tex
MDP algorithm in hand, we are now ready to provide an algorithm for
stochastic games.
%At a high level, this algorithm works by initially
%planning for $\pTwo$ selecting the action that minimizes randomness
%(with ties broken by performance). This assumption leads to a
%reduction to the MDP-case, where the rationality indexed family,
%$\sched_\rat$, indexes the Pareto front.  If this assumption is ever
%violated, the resulting state must support more randomness.  The
%rationality, $\rat$, is thus lowered to match the worst case randomness,
%which due to monotonicity of $\solfuncp$, can only increase
%performance. Surprisingly, as we shall later prove, this class of
%policies indexes the Pareto front for SGs!

\mypara{Environment Policies} We begin with three observations about
the $\pTwo$-policies.  First, for ERCI, we can assume an adversary for
$\pTwo$ that aims to foil $\pOne$ achieving both the performance
\emph{and} randomization requirement. We call such a $\pTwo$-policy
\emph{violating.} For a policy to be violating, it suffices to
violate, against every $\pOne$-policy independently, either
performance \emph{or} randomization.  Second, if there is a violating
$\pTwo$-policy, there is a deterministic $\pTwo$-policy that proves
this.  In particular, at every state, $\pTwoSched$ may choose to
violate either constraint via the appropriate action with no incentive
to randomize. Third, fixing an environment policy reduces $\sg$ to a
MDP $\sg[\pTwoSched]$.

\begin{figure}[t]
\centering
\scalebox{0.8}{
\input{sg_vis.tex}
}
\caption{SG to illustrate entropy matching policies.}
\label{fig:sg:simplest}
\end{figure}
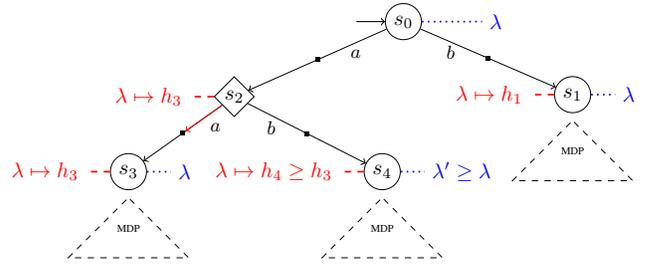

\mypara{A Sufficient Class of Policies}
For MDPs, we have seen that varying rationality is sufficient to explore the Pareto curve. 
We show that we can adapt that idea to a class we call \emph{entropy matching policies}, which may be indexed by the (initial) rationality. 
In the initial state, we start by assuming that $\pTwo$
selects a (deterministic) policy, $\sched^\rat_\pTwo$, that
lexicographically minimizes the guaranteed randomness, followed by
performance. On the sub-graph, $\sg[\sched^\rat_\pTwo]$, $\pOne$
employs the corresponding entropy maximizing policy for the MDP $\sg[\sched^\rat_\pTwo]$. 
Whenever $\pTwo$ diverges from the entropy minimizing policy (to decrease the induced performance), we let $\pOne$ increase its rationality such that it still induces the same guaranteed randomness.  We refer to this idea as \emph{entropy matching}. The idea is that the rationality at the initial state induces a worst-case entropy, and whatever $\pTwo$ chooses to do, throughout the SG, we ensure that we indeed obtain this entropy. 
The policy thus tracks this entropy and if necessary adapts the rationality (which we call \emph{replanning}). 
Replanning ensures we obtain the optimal performance from a particular point while still ensuring the required randomness.

\begin{example}
  We sketch an entropy matching policy in
  Fig.~\ref{fig:sg:simplest}. In particular, we show part of a SG. For
  some fixed rationality $\rat$, we annotate in red, on the left of
  the SG states, the entropy obtained when assuming that $\pTwo$ plays
  an entropy-minimizing policy as outlined above. In particular, this
  means that in $s_2$, $\pTwo$ selects action $a$. Now, our
  entropy-matching policy (in blue, on the right) will play with
  rationality $\rat$, unless state $s_4$ is reached. As this ensures a
  higher entropy, we may now select a higher rationality, $\rat'$.
\end{example}

%
%Our algorithm operates by picking an optimization direction
%$\langle \rat, 1 \rangle$ and \emph{temporarily} assuming that $\pTwo$
%selects a (deterministic) policy, $\sched^\rat_\pTwo$, that
%lexicographically minimizes the guaranteed randomness, followed by
%performance. On the sub-graph, $\sg[\sched^\rat_\pTwo]$, $\pOne$
%employs the corresponding entropy maximizing policy for the MDP,
%$\sg[\sched^\rat_\pTwo]$. This partial-policy is extended to a policy
%on the rest of the $\sg$ as follows. Whenever $\pTwo$ diverges from
%the entropy minimizing policy, it must be possible for $\pOne$ to
%trade randomness for performance, i.e. increase $\rat$, while still
%ensuring the same randomness against a entropy minimizing
%adversary.\sj{This needs at least one more sentence} Thus, a new MDP partial-policy\sj{???} is computed and extended
%recursively to all states of the SG.  We call the increase
%in rationality (and the associated computations) \emph{replanning} and
%the corresponding family of policies, $\{\sched^\rat_\pOne\}_\rat$,
%\emph{entropy matching}.  Finally, 

\mypara{Soundness and Completeness}
Importantly, observe that because
fixing a policy for $\pOne$ yields a verifiable point in $\solutions$, any witness
for realizability we find is trivially sound. For completeness, we can
restrict ourselves to the case in which our algorithm claims the ERCI
instance unrealizable. Surprisingly, the class of policies we consider
suffices, and the algorithm is thus sound and (whenever halting)
complete (proof provided in Sec~\ref{sec:proofs}). That is, all
guaranteed points are witnessed by an entropy matching policy!

Further, observe that as a corollary of the entropy matching family
being complete, it must be the case that $\solfuncp(\rndp_\rat)$
inherits continuity and (strict) monotonicity from the MDP
case. Namely, at each $\pTwo$ state, the achievable points
$\solutions$ are necessarily the intersection of the achievable points
of the sub-graphs. By induction, (with the MDP base case), we obtain
continuity and strict monotonicity.

\mypara{Algorithm: Memoizing Pareto Fronts}
We propose approximating the  Pareto front using the same three staged sequence of exploring 
rationality coefficients (at the initial state) as the MDP case: (1) endpoints, (2) doubling,
(3) binary search.

To perform the above computations efficiently, we adopt a geometric
perspective. Namely, observe that each node of $\sg$ indexes a
sub-graph, which has a corresponding Pareto front for trading
performance for randomness. Further, note that the Pareto front at an
$\pTwo$ node is the intersection of the Pareto fronts of its child
nodes. Entropy matching corresponds to ``switching'' between Pareto
fronts and adjusting the optimization direction by increasing the
rationality.  Thus, by traversing the graph from the terminal states
to the initial state, approximating Pareto fronts along the way, one
can memoize how to trade performance for randomness at any given
node. This preprocessing enables determining the minimum entropy
response for any optimization direction and quickly replanning via a
convex combination of Pareto optimal policies.

\mypara{Approximate Pareto Fronts}
Of course, by varying $\rat$, one can only construct approximate
Pareto fronts $\hat{\pareto{}} \subseteq \pareto{\solutions}$.
%, where we denote the downward closure of $\hat{\pareto{}}$ as $\hat{\solutions} \subseteq \solutions$.
% let $\hat{f}_\solutions^{s'}$ denote the characterising function.
We propose the following high-level algorithm to adapt the above
algorithm to the case where each Pareto front approximation introduces at
most $\kappa$ error along the performance axis.
\begin{mdframed}
\begin{enumerate}
\item Let $\tau$ denote the length of the longest path in $\sg$.
\item Let $0 < \kappa < 1$ be some arbitrary initial tolerance.
\item Recursively compute $\kappa$-close Pareto fronts for each successor state using replanning.
\item If the any minimum entropy action cannot be determined or $\scthreshold$ is within $\kappa\cdot \tau$ distance to (but outside of) $\hat{\pareto{}}$,
  halve $\kappa$ and repeat.
\item Otherwise, perform the entropy matching algorithm (with initial
  entropy $\randomness$) using these Pareto fronts and return the
  resulting policy (if on exists).
\end{enumerate}  
\end{mdframed}
The soundness of this algorithm relies on the following critical
facts: (1) Given sufficient resolution, the minimum entropy
$\pTwo$-actions can be determined. (2) The
resulting entropy depends solely on
the resulting sub-graph (and is independent of the current Pareto
approximation). (3) Thus, when querying points on
$\pareto{\solutions}$, error can only accumulate for $\scp$. (4) Next,
observe that $\scp$ is computed using convex combinations of entropy
matched points on Pareto approximations. (5) Convex combinations of an error interval cannot
increase the error, i.e.,
\begin{equation}
  q\cdot[x, x + \kappa] + \bar{q}\cdot[y, y + \kappa] = [z, z + \kappa],
\end{equation}
where $z = q\cdot x + \bar{q}\cdot y$.
Thus, so long as $\kappa\cdot\tau$ is enough resolution to answer $\scp_\rat <
\scthreshold$, one obtains a semi-decision procedure as in the MDP
case.

\mypara{Termination and Run Time} First, as in the MDP case, the
algorithm terminates almost surely, with the exception occurring only
for a subset of the Pareto front.  Below, we give an output-sensitive
analysis of the run time (assuming it does halt).  If $\kappa^*$
tolerance is required to terminate, then the $\kappa$ search
introduces $\mathcal{O}(\log(\nicefrac{1}{\kappa^*}))$
iterations. Next, observe that each node need process a given rationality
coefficient at most once. Further, looking up which pair of rationalities
are need to upper and lower bound the performance for a given randomness
can be done in logarithmic time via binary search on rationality coefficients.
As the corresponding bounds and convex combinations can be computed in
constant time, this means this algorithm runs in time:
\begin{equation}
  \label{eq:sg-runtime}
  \mathcal{O}\Big(\log(\nicefrac{1}{\kappa^*})\cdot N_\rat\cdot \log(N_\rat)\cdot |\sg|\Big),
\end{equation}
where, $N_\rat$ is the number of unique rationality coefficients
processed.  If, as in the MDP case, one assumes a maximum rationality
coefficient $\rat^*$ and a minimum rationality resolution $\Delta$,
one obtains:
\begin{equation}
  \mathcal{O}\Big(\underbrace{\log(\nicefrac{1}{\kappa^*})}_{\kappa \text{ search}}\cdot \underbrace{\nicefrac{\rat^*}{\Delta}\cdot \overbrace{\log(\nicefrac{\rat^*}{\Delta})}^{\text{Replanning}}\cdot |\sg|}_{\text{Evaluate } \rat}\Big).
\end{equation}
The above however is very conservative and empirically we observe
$N_\rat$ bounded far away from $\nicefrac{\rat^*}{\Delta}$.

%In practice, this algorithm can be significantly improved by adaptive
%tolerances, lazily computing the Pareto Fronts, and only computing
%Pareto Fronts for $\pTwo$ states. Nevertheless,
%already this na\"ive algorithm gives a sense of the run-time
%bottlenecks. Namely, if $\kappa^*$ tolerance is required to terminate,
%then the $\kappa$ search introduces $O(\log(\nicefrac{1}{\kappa^*}))$
%iterations. Furthermore, by computing $O(|\sg|)$ Pareto fronts, from the
%leaves, one ensures that the complexity grows linearly with the graph
%size - although the multiplicative constant depends on the number of
%rationality coefficients explored per Pareto front. We found that most
%of the rationality coefficients explored were shared, which in
%practice seems to amortize the cost per state. Finally, as in the
%MDP-case, the algorithm halts if $\langle \scthreshold, \randomness \rangle$ is
%bounded away from the root Pareto Front. As the Pareto Front has
%measure 0, we argue that this is merely a technical concern, as a
%small perturbation to the ERCI instance (i.e. a Smoothed
%Analysis~\cite{SmoothedAnalysis}) on $\sg$ admits decidability.

%%% Local Variables:
%%% mode: latex
%%% TeX-master: "main"
%%% End:

%% file: sg_vis.tex
\begin{tikzpicture}
    \node[sstate, initial, initial text=,initial where=left] (a1) {$s_0$}; 
    
    \node[right=of a1,color=blue] (a1r) {$\lambda$ }; 
    \draw[dotted,blue, thick] (a1) -- (a1r);
    
	\node[sstate,below=0.6cm of a1,xshift=8em] (a2) {$s_1$};
	
	\node[astate,below=0.6cm of a1,xshift=-8em] (a0) {$s_2$};
	\node[sstate,below=0.6cm of a0,xshift=-5em] (s1) {$s_3$};
	\node[sstate,below=0.6cm  of a0,xshift=7em] (s2) {$s_4$};

	\node[below=0.1cm of s1, inner sep=0.3pt] (x1) {};
	
	\node[below=0.1cm of s2, inner sep=0.3pt] (x2) {};
		\node[below=0.1cm of a2, inner sep=0.3pt] (x3) {};
	
	\draw[->] (a0) -- node[actnode] (x) {}
					  node[pos=0.4,elab,right,xshift=2mm] {$a$}  (s1);
	\draw[->] (a0) -- node[actnode] {}
					  node[pos=0.4,elab,left,xshift=-2mm] {$b$}  (s2);
					  
	\draw[->] (a1) -- node[actnode] {}
					  node[pos=0.4,elab,right,xshift=2mm] {$a$}  (a0);
	\draw[->] (a1) -- node[actnode] {}
					  node[pos=0.4,elab,left,xshift=-2mm] {$b$}  (a2);
					  
	\draw[->, red] (a0) -- (x);
	
	\node[right=0.4cm of s1,color=blue] (s3r) {$\lambda$ }; 
    \draw[dotted,blue, thick] (s1) -- (s3r);
    \node[right=0.4cm of s2,color=blue] (s4r) {$\lambda' \geq \lambda $}; 
    \draw[dotted,blue, thick] (s2) -- (s4r);
    \node[right=0.4cm of a2,color=blue] (a2r) {$\lambda$ }; 
    \draw[dotted,blue, thick] (a2) -- (a2r);
    \node[left=0.4cm of s1,color=red] (s3r) {$\lambda \mapsto h_3$ }; 
    \draw[dashed,red, thick] (s1) -- (s3r);
    \node[left=0.4cm of s2,color=red] (s4r) {$\lambda \mapsto h_4 \geq h_3$}; 
    \draw[dashed,red, thick] (s2) -- (s4r);
    \node[left=0.4cm of a2,color=red] (a2r) {$\lambda \mapsto h_1$ }; 
    \draw[dashed,red, thick] (a2) -- (a2r);
    \node[left=0.4cm of a0,color=red] (s2r) {$\lambda \mapsto h_3$ }; 
    \draw[dashed,red, thick] (a0) -- (s2r);

	\draw[dashed] (x1) -- +(1,-1) -- +(-1,-1) coordinate  (fa) -- (x1);
	\draw[dashed] (x2) -- +(1,-1) coordinate (fb) -- +(-1,-1)  -- (x2);
	
	\draw[dashed] (x3) -- +(1,-1) -- +(-1,-1)  -- (x3);
	
	\node[below=3mm of x1] {{\tiny MDP}};
	\node[below=3mm of x2] {{\tiny MDP}};
	\node[below=3mm of x3] {{\tiny MDP}};
	
%	\node[fit=(a0)(x1)(x2)(fa)(fb), inner sep = 6pt, dotted,draw] {};

\end{tikzpicture}

%%% Local Variables:
%%% mode: latex
%%% TeX-master: "main"
%%% End:

%% file: experiment.tex
\begin{figure*}
    \begin{subfigure}{0.47\textwidth}
  \centering
  \scalebox{0.50}{
    \input{imgs/pareto_solve_time.pgf}
    }
    \caption{
      Experimental times for computing Pareto front of a variety
      delivery drone problems\label{fig:exp_times}.
    }
  \end{subfigure}
  \hfill
    \begin{subfigure}{0.47\textwidth}
    \centering \scalebox{0.53}{
      \input{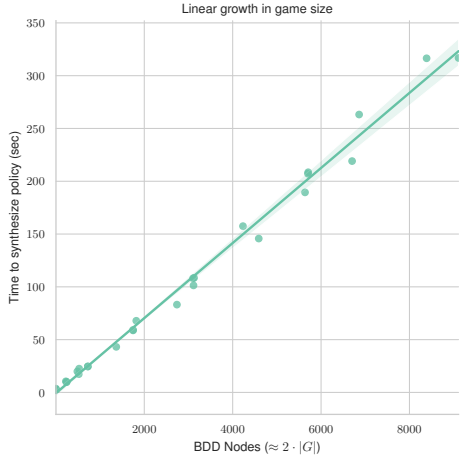}
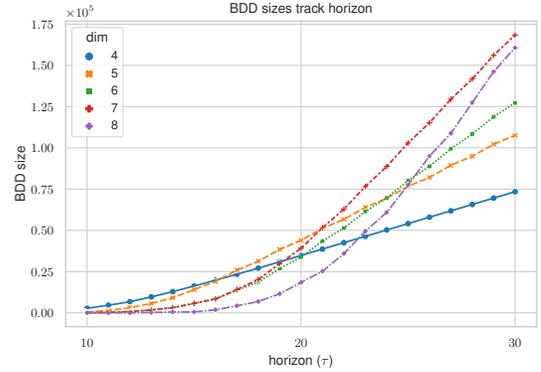
    }
    \caption{
      BDD graph size as a function of horizon for the problems in our
      benchmark suite. Distribution of problems, non-uniform in horizon
      to avoid small horizon artifacts\label{fig:bdd_sizes}.
    }
  \end{subfigure}
  \caption{Plots to illustrate scalability} 
\end{figure*}

\mypara{Setup} To experimentally validate the feasibility of our ERCI
algorithm for SGs, we implemented~\cite{RSS21code} our algorithm in
Python.  Inspired by the recent work on compressing MDPs for
specification inference~\cite{DBLP:conf/cav/Vazquez-Chanlatte20}, each
SG was represented as a Binary Decision Diagram
(BDD)~\cite{DBLP:journals/csur/Bryant92} using the \texttt{dd} and
\texttt{py-aiger} python packages~\cite{dd, pyAiger}.

We investigate the motivating example.
Specifically, our experiments used
a $k\times k$ grid discretization of the workspace (cf.\ Fig.~\ref{fig:motivating}), for $k \in
\{4,5,6,7\}$ where the four target houses lie in
$\{\lfloor\nicefrac{k}{3}\rfloor, \lfloor\nicefrac{2k}{3}\rfloor
\}^2$, and the drones $\droneEgo$ and $\droneEnv$ are initially at in the bottom left corner and top right house resp. Furthermore, for simplicity, we embedded the avoid crash
condition as part of the soft constraint, rather than a hard
constraint\footnotemark. We took~$\pOne$'s dynamics to be deterministic
and modeled $\pTwo$ as visiting each house in either clock-wise or
counter-wise order, where the orientation can switch with
(a potentially state dependent) probability $p \in [\nicefrac{1}{100},
\nicefrac{1}{50}]$ whenever a house is visited. Next, we considered an
alternation between $\pOne$ and $\pTwo$ to be a single logical time
step, and (non-uniformly) instantiate problem instances with horizons
ranging from $6$ to $18$, i.e., paths ranged from length $12$ to
$36$.
\footnotetext{Note that counter-intuitively, only using soft
constraints generally results in harder instances as the compressed SGs are larger.}

\mypara{Results} First and foremost, we succeed in synthesizing
controllers in the mentioned setup.  The controller randomizes its
behavior while meeting the specification, which is not surprising as
the algorithm yields a correct-by-construction policy.

Next, we consider the practical run time of our algorithm.  As
Fig~\ref{fig:exp_times} demonstrates, the empirical time to estimate
the Pareto front seemed to increase linearly with our SG 
encoding -- which is consistent with our complexity analysis.
Moreover, our encoding seems to linearly track with the horizon for
all $k$ (Fig.~\ref{fig:bdd_sizes}), suggesting that the overall run
time grows linearly in the horizon within our parameterization. When
combined with the potential to parallelize across the rationality
coefficients, these results suggest that practical optimizations to
our ERCI algorithm may admit usage on other more complicated
benchmarks. Finally, we remark that the use of a decision diagram
encoding did indeed dramatically decrease the size of the SG 
(with negligible overhead).\footnotemark

\footnotetext{For example, the $(k=8,\text{horizon}=18)$ case is
encoded using a 505,100 node BDD ($|\sg| = 6,861$ nodes). Compare with
the direct encoding $|\sg| = |S|\cdot \tau \cdot |\text{monitor state}| = (8\times 8)^2 \cdot (2 \cdot 18) \cdot (2^4\cdot 2) \approx 37,000$.}

%%% Local Variables:
%%% mode: latex
%%% TeX-master: "main"
%%% End:

%% file: imgs/pareto_solve_time.pgf
%% Creator: Matplotlib, PGF backend
%%
%% To include the figure in your LaTeX document, write
%%   \input{<filename>.pgf}
%%
%% Make sure the required packages are loaded in your preamble
%%   \usepackage{pgf}
%%
%% and, on pdftex
%%   \usepackage[utf8]{inputenc}\DeclareUnicodeCharacter{2212}{-}
%%
%% or, on luatex and xetex
%%   \usepackage{unicode-math}
%%
%% Figures using additional raster images can only be included by \input if
%% they are in the same directory as the main LaTeX file. For loading figures
%% from other directories you can use the `import` package
%% and then include the figures with
%%   \import{<path to file>}{<filename>.pgf}
%%
%% Matplotlib used the following preamble
%%
\begingroup%
\makeatletter%
\begin{pgfpicture}%
\pgfpathrectangle{\pgfpointorigin}{\pgfqpoint{5.000000in}{5.000000in}}%
\pgfusepath{use as bounding box, clip}%
\begin{pgfscope}%
\pgfsetbuttcap%
\pgfsetmiterjoin%
\definecolor{currentfill}{rgb}{1.000000,1.000000,1.000000}%
\pgfsetfillcolor{currentfill}%
\pgfsetlinewidth{0.000000pt}%
\definecolor{currentstroke}{rgb}{1.000000,1.000000,1.000000}%
\pgfsetstrokecolor{currentstroke}%
\pgfsetstrokeopacity{0.000000}%
\pgfsetdash{}{0pt}%
\pgfpathmoveto{\pgfqpoint{0.000000in}{0.000000in}}%
\pgfpathlineto{\pgfqpoint{5.000000in}{0.000000in}}%
\pgfpathlineto{\pgfqpoint{5.000000in}{5.000000in}}%
\pgfpathlineto{\pgfqpoint{0.000000in}{5.000000in}}%
\pgfpathclose%
\pgfusepath{fill}%
\end{pgfscope}%
\begin{pgfscope}%
\pgfsetbuttcap%
\pgfsetmiterjoin%
\definecolor{currentfill}{rgb}{1.000000,1.000000,1.000000}%
\pgfsetfillcolor{currentfill}%
\pgfsetlinewidth{0.000000pt}%
\definecolor{currentstroke}{rgb}{0.000000,0.000000,0.000000}%
\pgfsetstrokecolor{currentstroke}%
\pgfsetstrokeopacity{0.000000}%
\pgfsetdash{}{0pt}%
\pgfpathmoveto{\pgfqpoint{0.641735in}{0.555904in}}%
\pgfpathlineto{\pgfqpoint{4.856000in}{0.555904in}}%
\pgfpathlineto{\pgfqpoint{4.856000in}{4.680420in}}%
\pgfpathlineto{\pgfqpoint{0.641735in}{4.680420in}}%
\pgfpathclose%
\pgfusepath{fill}%
\end{pgfscope}%
\begin{pgfscope}%
\pgfpathrectangle{\pgfqpoint{0.641735in}{0.555904in}}{\pgfqpoint{4.214265in}{4.124515in}}%
\pgfusepath{clip}%
\pgfsetroundcap%
\pgfsetroundjoin%
\pgfsetlinewidth{0.803000pt}%
\definecolor{currentstroke}{rgb}{0.800000,0.800000,0.800000}%
\pgfsetstrokecolor{currentstroke}%
\pgfsetdash{}{0pt}%
\pgfpathmoveto{\pgfqpoint{1.566107in}{0.555904in}}%
\pgfpathlineto{\pgfqpoint{1.566107in}{4.680420in}}%
\pgfusepath{stroke}%
\end{pgfscope}%
\begin{pgfscope}%
\definecolor{textcolor}{rgb}{0.150000,0.150000,0.150000}%
\pgfsetstrokecolor{textcolor}%
\pgfsetfillcolor{textcolor}%
\pgftext[x=1.566107in,y=0.440627in,,top]{\color{textcolor}\sffamily\fontsize{8.800000}{10.560000}\selectfont \(\displaystyle {2000}\)}%
\end{pgfscope}%
\begin{pgfscope}%
\pgfpathrectangle{\pgfqpoint{0.641735in}{0.555904in}}{\pgfqpoint{4.214265in}{4.124515in}}%
\pgfusepath{clip}%
\pgfsetroundcap%
\pgfsetroundjoin%
\pgfsetlinewidth{0.803000pt}%
\definecolor{currentstroke}{rgb}{0.800000,0.800000,0.800000}%
\pgfsetstrokecolor{currentstroke}%
\pgfsetdash{}{0pt}%
\pgfpathmoveto{\pgfqpoint{2.491404in}{0.555904in}}%
\pgfpathlineto{\pgfqpoint{2.491404in}{4.680420in}}%
\pgfusepath{stroke}%
\end{pgfscope}%
\begin{pgfscope}%
\definecolor{textcolor}{rgb}{0.150000,0.150000,0.150000}%
\pgfsetstrokecolor{textcolor}%
\pgfsetfillcolor{textcolor}%
\pgftext[x=2.491404in,y=0.440627in,,top]{\color{textcolor}\sffamily\fontsize{8.800000}{10.560000}\selectfont \(\displaystyle {4000}\)}%
\end{pgfscope}%
\begin{pgfscope}%
\pgfpathrectangle{\pgfqpoint{0.641735in}{0.555904in}}{\pgfqpoint{4.214265in}{4.124515in}}%
\pgfusepath{clip}%
\pgfsetroundcap%
\pgfsetroundjoin%
\pgfsetlinewidth{0.803000pt}%
\definecolor{currentstroke}{rgb}{0.800000,0.800000,0.800000}%
\pgfsetstrokecolor{currentstroke}%
\pgfsetdash{}{0pt}%
\pgfpathmoveto{\pgfqpoint{3.416701in}{0.555904in}}%
\pgfpathlineto{\pgfqpoint{3.416701in}{4.680420in}}%
\pgfusepath{stroke}%
\end{pgfscope}%
\begin{pgfscope}%
\definecolor{textcolor}{rgb}{0.150000,0.150000,0.150000}%
\pgfsetstrokecolor{textcolor}%
\pgfsetfillcolor{textcolor}%
\pgftext[x=3.416701in,y=0.440627in,,top]{\color{textcolor}\sffamily\fontsize{8.800000}{10.560000}\selectfont \(\displaystyle {6000}\)}%
\end{pgfscope}%
\begin{pgfscope}%
\pgfpathrectangle{\pgfqpoint{0.641735in}{0.555904in}}{\pgfqpoint{4.214265in}{4.124515in}}%
\pgfusepath{clip}%
\pgfsetroundcap%
\pgfsetroundjoin%
\pgfsetlinewidth{0.803000pt}%
\definecolor{currentstroke}{rgb}{0.800000,0.800000,0.800000}%
\pgfsetstrokecolor{currentstroke}%
\pgfsetdash{}{0pt}%
\pgfpathmoveto{\pgfqpoint{4.341998in}{0.555904in}}%
\pgfpathlineto{\pgfqpoint{4.341998in}{4.680420in}}%
\pgfusepath{stroke}%
\end{pgfscope}%
\begin{pgfscope}%
\definecolor{textcolor}{rgb}{0.150000,0.150000,0.150000}%
\pgfsetstrokecolor{textcolor}%
\pgfsetfillcolor{textcolor}%
\pgftext[x=4.341998in,y=0.440627in,,top]{\color{textcolor}\sffamily\fontsize{8.800000}{10.560000}\selectfont \(\displaystyle {8000}\)}%
\end{pgfscope}%
\begin{pgfscope}%
\definecolor{textcolor}{rgb}{0.150000,0.150000,0.150000}%
\pgfsetstrokecolor{textcolor}%
\pgfsetfillcolor{textcolor}%
\pgftext[x=2.748867in,y=0.273960in,,top]{\color{textcolor}\sffamily\fontsize{9.600000}{11.520000}\selectfont BDD Nodes (\(\displaystyle \approx 2 \cdot |G|\))}%
\end{pgfscope}%
\begin{pgfscope}%
\pgfpathrectangle{\pgfqpoint{0.641735in}{0.555904in}}{\pgfqpoint{4.214265in}{4.124515in}}%
\pgfusepath{clip}%
\pgfsetroundcap%
\pgfsetroundjoin%
\pgfsetlinewidth{0.803000pt}%
\definecolor{currentstroke}{rgb}{0.800000,0.800000,0.800000}%
\pgfsetstrokecolor{currentstroke}%
\pgfsetdash{}{0pt}%
\pgfpathmoveto{\pgfqpoint{0.641735in}{0.785980in}}%
\pgfpathlineto{\pgfqpoint{4.856000in}{0.785980in}}%
\pgfusepath{stroke}%
\end{pgfscope}%
\begin{pgfscope}%
\definecolor{textcolor}{rgb}{0.150000,0.150000,0.150000}%
\pgfsetstrokecolor{textcolor}%
\pgfsetfillcolor{textcolor}%
\pgftext[x=0.462222in, y=0.742577in, left, base]{\color{textcolor}\sffamily\fontsize{8.800000}{10.560000}\selectfont \(\displaystyle {0}\)}%
\end{pgfscope}%
\begin{pgfscope}%
\pgfpathrectangle{\pgfqpoint{0.641735in}{0.555904in}}{\pgfqpoint{4.214265in}{4.124515in}}%
\pgfusepath{clip}%
\pgfsetroundcap%
\pgfsetroundjoin%
\pgfsetlinewidth{0.803000pt}%
\definecolor{currentstroke}{rgb}{0.800000,0.800000,0.800000}%
\pgfsetstrokecolor{currentstroke}%
\pgfsetdash{}{0pt}%
\pgfpathmoveto{\pgfqpoint{0.641735in}{1.339030in}}%
\pgfpathlineto{\pgfqpoint{4.856000in}{1.339030in}}%
\pgfusepath{stroke}%
\end{pgfscope}%
\begin{pgfscope}%
\definecolor{textcolor}{rgb}{0.150000,0.150000,0.150000}%
\pgfsetstrokecolor{textcolor}%
\pgfsetfillcolor{textcolor}%
\pgftext[x=0.397986in, y=1.295628in, left, base]{\color{textcolor}\sffamily\fontsize{8.800000}{10.560000}\selectfont \(\displaystyle {50}\)}%
\end{pgfscope}%
\begin{pgfscope}%
\pgfpathrectangle{\pgfqpoint{0.641735in}{0.555904in}}{\pgfqpoint{4.214265in}{4.124515in}}%
\pgfusepath{clip}%
\pgfsetroundcap%
\pgfsetroundjoin%
\pgfsetlinewidth{0.803000pt}%
\definecolor{currentstroke}{rgb}{0.800000,0.800000,0.800000}%
\pgfsetstrokecolor{currentstroke}%
\pgfsetdash{}{0pt}%
\pgfpathmoveto{\pgfqpoint{0.641735in}{1.892081in}}%
\pgfpathlineto{\pgfqpoint{4.856000in}{1.892081in}}%
\pgfusepath{stroke}%
\end{pgfscope}%
\begin{pgfscope}%
\definecolor{textcolor}{rgb}{0.150000,0.150000,0.150000}%
\pgfsetstrokecolor{textcolor}%
\pgfsetfillcolor{textcolor}%
\pgftext[x=0.333750in, y=1.848678in, left, base]{\color{textcolor}\sffamily\fontsize{8.800000}{10.560000}\selectfont \(\displaystyle {100}\)}%
\end{pgfscope}%
\begin{pgfscope}%
\pgfpathrectangle{\pgfqpoint{0.641735in}{0.555904in}}{\pgfqpoint{4.214265in}{4.124515in}}%
\pgfusepath{clip}%
\pgfsetroundcap%
\pgfsetroundjoin%
\pgfsetlinewidth{0.803000pt}%
\definecolor{currentstroke}{rgb}{0.800000,0.800000,0.800000}%
\pgfsetstrokecolor{currentstroke}%
\pgfsetdash{}{0pt}%
\pgfpathmoveto{\pgfqpoint{0.641735in}{2.445132in}}%
\pgfpathlineto{\pgfqpoint{4.856000in}{2.445132in}}%
\pgfusepath{stroke}%
\end{pgfscope}%
\begin{pgfscope}%
\definecolor{textcolor}{rgb}{0.150000,0.150000,0.150000}%
\pgfsetstrokecolor{textcolor}%
\pgfsetfillcolor{textcolor}%
\pgftext[x=0.333750in, y=2.401729in, left, base]{\color{textcolor}\sffamily\fontsize{8.800000}{10.560000}\selectfont \(\displaystyle {150}\)}%
\end{pgfscope}%
\begin{pgfscope}%
\pgfpathrectangle{\pgfqpoint{0.641735in}{0.555904in}}{\pgfqpoint{4.214265in}{4.124515in}}%
\pgfusepath{clip}%
\pgfsetroundcap%
\pgfsetroundjoin%
\pgfsetlinewidth{0.803000pt}%
\definecolor{currentstroke}{rgb}{0.800000,0.800000,0.800000}%
\pgfsetstrokecolor{currentstroke}%
\pgfsetdash{}{0pt}%
\pgfpathmoveto{\pgfqpoint{0.641735in}{2.998182in}}%
\pgfpathlineto{\pgfqpoint{4.856000in}{2.998182in}}%
\pgfusepath{stroke}%
\end{pgfscope}%
\begin{pgfscope}%
\definecolor{textcolor}{rgb}{0.150000,0.150000,0.150000}%
\pgfsetstrokecolor{textcolor}%
\pgfsetfillcolor{textcolor}%
\pgftext[x=0.333750in, y=2.954780in, left, base]{\color{textcolor}\sffamily\fontsize{8.800000}{10.560000}\selectfont \(\displaystyle {200}\)}%
\end{pgfscope}%
\begin{pgfscope}%
\pgfpathrectangle{\pgfqpoint{0.641735in}{0.555904in}}{\pgfqpoint{4.214265in}{4.124515in}}%
\pgfusepath{clip}%
\pgfsetroundcap%
\pgfsetroundjoin%
\pgfsetlinewidth{0.803000pt}%
\definecolor{currentstroke}{rgb}{0.800000,0.800000,0.800000}%
\pgfsetstrokecolor{currentstroke}%
\pgfsetdash{}{0pt}%
\pgfpathmoveto{\pgfqpoint{0.641735in}{3.551233in}}%
\pgfpathlineto{\pgfqpoint{4.856000in}{3.551233in}}%
\pgfusepath{stroke}%
\end{pgfscope}%
\begin{pgfscope}%
\definecolor{textcolor}{rgb}{0.150000,0.150000,0.150000}%
\pgfsetstrokecolor{textcolor}%
\pgfsetfillcolor{textcolor}%
\pgftext[x=0.333750in, y=3.507830in, left, base]{\color{textcolor}\sffamily\fontsize{8.800000}{10.560000}\selectfont \(\displaystyle {250}\)}%
\end{pgfscope}%
\begin{pgfscope}%
\pgfpathrectangle{\pgfqpoint{0.641735in}{0.555904in}}{\pgfqpoint{4.214265in}{4.124515in}}%
\pgfusepath{clip}%
\pgfsetroundcap%
\pgfsetroundjoin%
\pgfsetlinewidth{0.803000pt}%
\definecolor{currentstroke}{rgb}{0.800000,0.800000,0.800000}%
\pgfsetstrokecolor{currentstroke}%
\pgfsetdash{}{0pt}%
\pgfpathmoveto{\pgfqpoint{0.641735in}{4.104284in}}%
\pgfpathlineto{\pgfqpoint{4.856000in}{4.104284in}}%
\pgfusepath{stroke}%
\end{pgfscope}%
\begin{pgfscope}%
\definecolor{textcolor}{rgb}{0.150000,0.150000,0.150000}%
\pgfsetstrokecolor{textcolor}%
\pgfsetfillcolor{textcolor}%
\pgftext[x=0.333750in, y=4.060881in, left, base]{\color{textcolor}\sffamily\fontsize{8.800000}{10.560000}\selectfont \(\displaystyle {300}\)}%
\end{pgfscope}%
\begin{pgfscope}%
\pgfpathrectangle{\pgfqpoint{0.641735in}{0.555904in}}{\pgfqpoint{4.214265in}{4.124515in}}%
\pgfusepath{clip}%
\pgfsetroundcap%
\pgfsetroundjoin%
\pgfsetlinewidth{0.803000pt}%
\definecolor{currentstroke}{rgb}{0.800000,0.800000,0.800000}%
\pgfsetstrokecolor{currentstroke}%
\pgfsetdash{}{0pt}%
\pgfpathmoveto{\pgfqpoint{0.641735in}{4.657334in}}%
\pgfpathlineto{\pgfqpoint{4.856000in}{4.657334in}}%
\pgfusepath{stroke}%
\end{pgfscope}%
\begin{pgfscope}%
\definecolor{textcolor}{rgb}{0.150000,0.150000,0.150000}%
\pgfsetstrokecolor{textcolor}%
\pgfsetfillcolor{textcolor}%
\pgftext[x=0.333750in, y=4.613932in, left, base]{\color{textcolor}\sffamily\fontsize{8.800000}{10.560000}\selectfont \(\displaystyle {350}\)}%
\end{pgfscope}%
\begin{pgfscope}%
\definecolor{textcolor}{rgb}{0.150000,0.150000,0.150000}%
\pgfsetstrokecolor{textcolor}%
\pgfsetfillcolor{textcolor}%
\pgftext[x=0.278195in,y=2.618162in,,bottom,rotate=90.000000]{\color{textcolor}\sffamily\fontsize{9.600000}{11.520000}\selectfont Time to synthesize policy (sec)}%
\end{pgfscope}%
\begin{pgfscope}%
\pgfpathrectangle{\pgfqpoint{0.641735in}{0.555904in}}{\pgfqpoint{4.214265in}{4.124515in}}%
\pgfusepath{clip}%
\pgfsetbuttcap%
\pgfsetroundjoin%
\definecolor{currentfill}{rgb}{0.400000,0.760784,0.647059}%
\pgfsetfillcolor{currentfill}%
\pgfsetfillopacity{0.800000}%
\pgfsetlinewidth{1.003750pt}%
\definecolor{currentstroke}{rgb}{0.400000,0.760784,0.647059}%
\pgfsetstrokecolor{currentstroke}%
\pgfsetstrokeopacity{0.800000}%
\pgfsetdash{}{0pt}%
\pgfsys@defobject{currentmarker}{\pgfqpoint{-0.033333in}{-0.033333in}}{\pgfqpoint{0.033333in}{0.033333in}}{%
\pgfpathmoveto{\pgfqpoint{0.000000in}{-0.033333in}}%
\pgfpathcurveto{\pgfqpoint{0.008840in}{-0.033333in}}{\pgfqpoint{0.017319in}{-0.029821in}}{\pgfqpoint{0.023570in}{-0.023570in}}%
\pgfpathcurveto{\pgfqpoint{0.029821in}{-0.017319in}}{\pgfqpoint{0.033333in}{-0.008840in}}{\pgfqpoint{0.033333in}{0.000000in}}%
\pgfpathcurveto{\pgfqpoint{0.033333in}{0.008840in}}{\pgfqpoint{0.029821in}{0.017319in}}{\pgfqpoint{0.023570in}{0.023570in}}%
\pgfpathcurveto{\pgfqpoint{0.017319in}{0.029821in}}{\pgfqpoint{0.008840in}{0.033333in}}{\pgfqpoint{0.000000in}{0.033333in}}%
\pgfpathcurveto{\pgfqpoint{-0.008840in}{0.033333in}}{\pgfqpoint{-0.017319in}{0.029821in}}{\pgfqpoint{-0.023570in}{0.023570in}}%
\pgfpathcurveto{\pgfqpoint{-0.029821in}{0.017319in}}{\pgfqpoint{-0.033333in}{0.008840in}}{\pgfqpoint{-0.033333in}{0.000000in}}%
\pgfpathcurveto{\pgfqpoint{-0.033333in}{-0.008840in}}{\pgfqpoint{-0.029821in}{-0.017319in}}{\pgfqpoint{-0.023570in}{-0.023570in}}%
\pgfpathcurveto{\pgfqpoint{-0.017319in}{-0.029821in}}{\pgfqpoint{-0.008840in}{-0.033333in}}{\pgfqpoint{0.000000in}{-0.033333in}}%
\pgfpathclose%
\pgfusepath{stroke,fill}%
}%
\begin{pgfscope}%
\pgfsys@transformshift{1.908004in}{1.705730in}%
\pgfsys@useobject{currentmarker}{}%
\end{pgfscope}%
\begin{pgfscope}%
\pgfsys@transformshift{2.764366in}{2.397904in}%
\pgfsys@useobject{currentmarker}{}%
\end{pgfscope}%
\begin{pgfscope}%
\pgfsys@transformshift{3.741480in}{3.209554in}%
\pgfsys@useobject{currentmarker}{}%
\end{pgfscope}%
\begin{pgfscope}%
\pgfsys@transformshift{0.880462in}{0.977186in}%
\pgfsys@useobject{currentmarker}{}%
\end{pgfscope}%
\begin{pgfscope}%
\pgfsys@transformshift{1.272788in}{1.263379in}%
\pgfsys@useobject{currentmarker}{}%
\end{pgfscope}%
\begin{pgfscope}%
\pgfsys@transformshift{2.081497in}{1.907102in}%
\pgfsys@useobject{currentmarker}{}%
\end{pgfscope}%
\begin{pgfscope}%
\pgfsys@transformshift{3.248759in}{2.881560in}%
\pgfsys@useobject{currentmarker}{}%
\end{pgfscope}%
\begin{pgfscope}%
\pgfsys@transformshift{4.856000in}{4.289814in}%
\pgfsys@useobject{currentmarker}{}%
\end{pgfscope}%
\begin{pgfscope}%
\pgfsys@transformshift{0.641735in}{0.815623in}%
\pgfsys@useobject{currentmarker}{}%
\end{pgfscope}%
\begin{pgfscope}%
\pgfsys@transformshift{0.753233in}{0.893495in}%
\pgfsys@useobject{currentmarker}{}%
\end{pgfscope}%
\begin{pgfscope}%
\pgfsys@transformshift{0.975305in}{1.056484in}%
\pgfsys@useobject{currentmarker}{}%
\end{pgfscope}%
\begin{pgfscope}%
\pgfsys@transformshift{1.448594in}{1.435887in}%
\pgfsys@useobject{currentmarker}{}%
\end{pgfscope}%
\begin{pgfscope}%
\pgfsys@transformshift{2.087511in}{1.986288in}%
\pgfsys@useobject{currentmarker}{}%
\end{pgfscope}%
\begin{pgfscope}%
\pgfsys@transformshift{3.279294in}{3.077419in}%
\pgfsys@useobject{currentmarker}{}%
\end{pgfscope}%
\begin{pgfscope}%
\pgfsys@transformshift{0.641735in}{0.818967in}%
\pgfsys@useobject{currentmarker}{}%
\end{pgfscope}%
\begin{pgfscope}%
\pgfsys@transformshift{0.753233in}{0.895706in}%
\pgfsys@useobject{currentmarker}{}%
\end{pgfscope}%
\begin{pgfscope}%
\pgfsys@transformshift{0.975305in}{1.059266in}%
\pgfsys@useobject{currentmarker}{}%
\end{pgfscope}%
\begin{pgfscope}%
\pgfsys@transformshift{1.448594in}{1.440893in}%
\pgfsys@useobject{currentmarker}{}%
\end{pgfscope}%
\begin{pgfscope}%
\pgfsys@transformshift{2.076408in}{1.981670in}%
\pgfsys@useobject{currentmarker}{}%
\end{pgfscope}%
\begin{pgfscope}%
\pgfsys@transformshift{3.282070in}{3.092604in}%
\pgfsys@useobject{currentmarker}{}%
\end{pgfscope}%
\begin{pgfscope}%
\pgfsys@transformshift{4.521505in}{4.286374in}%
\pgfsys@useobject{currentmarker}{}%
\end{pgfscope}%
\begin{pgfscope}%
\pgfsys@transformshift{0.641735in}{0.825554in}%
\pgfsys@useobject{currentmarker}{}%
\end{pgfscope}%
\begin{pgfscope}%
\pgfsys@transformshift{0.745368in}{0.902267in}%
\pgfsys@useobject{currentmarker}{}%
\end{pgfscope}%
\begin{pgfscope}%
\pgfsys@transformshift{0.865657in}{1.006042in}%
\pgfsys@useobject{currentmarker}{}%
\end{pgfscope}%
\begin{pgfscope}%
\pgfsys@transformshift{0.884163in}{1.035157in}%
\pgfsys@useobject{currentmarker}{}%
\end{pgfscope}%
\begin{pgfscope}%
\pgfsys@transformshift{1.482367in}{1.536681in}%
\pgfsys@useobject{currentmarker}{}%
\end{pgfscope}%
\begin{pgfscope}%
\pgfsys@transformshift{2.599201in}{2.528121in}%
\pgfsys@useobject{currentmarker}{}%
\end{pgfscope}%
\begin{pgfscope}%
\pgfsys@transformshift{3.815041in}{3.697210in}%
\pgfsys@useobject{currentmarker}{}%
\end{pgfscope}%
\end{pgfscope}%
\begin{pgfscope}%
\pgfpathrectangle{\pgfqpoint{0.641735in}{0.555904in}}{\pgfqpoint{4.214265in}{4.124515in}}%
\pgfusepath{clip}%
\pgfsetbuttcap%
\pgfsetroundjoin%
\definecolor{currentfill}{rgb}{0.400000,0.760784,0.647059}%
\pgfsetfillcolor{currentfill}%
\pgfsetfillopacity{0.150000}%
\pgfsetlinewidth{0.803000pt}%
\definecolor{currentstroke}{rgb}{1.000000,1.000000,1.000000}%
\pgfsetstrokecolor{currentstroke}%
\pgfsetstrokeopacity{0.150000}%
\pgfsetdash{}{0pt}%
\pgfsys@defobject{currentmarker}{\pgfqpoint{0.641735in}{0.743382in}}{\pgfqpoint{4.856000in}{4.492942in}}{%
\pgfpathmoveto{\pgfqpoint{0.641735in}{0.806991in}}%
\pgfpathlineto{\pgfqpoint{0.641735in}{0.743382in}}%
\pgfpathlineto{\pgfqpoint{0.684303in}{0.780482in}}%
\pgfpathlineto{\pgfqpoint{0.726872in}{0.817839in}}%
\pgfpathlineto{\pgfqpoint{0.769440in}{0.855266in}}%
\pgfpathlineto{\pgfqpoint{0.812008in}{0.892481in}}%
\pgfpathlineto{\pgfqpoint{0.854577in}{0.929970in}}%
\pgfpathlineto{\pgfqpoint{0.897145in}{0.967549in}}%
\pgfpathlineto{\pgfqpoint{0.939713in}{1.004747in}}%
\pgfpathlineto{\pgfqpoint{0.982282in}{1.041810in}}%
\pgfpathlineto{\pgfqpoint{1.024850in}{1.078716in}}%
\pgfpathlineto{\pgfqpoint{1.067418in}{1.115674in}}%
\pgfpathlineto{\pgfqpoint{1.109987in}{1.152578in}}%
\pgfpathlineto{\pgfqpoint{1.152555in}{1.189538in}}%
\pgfpathlineto{\pgfqpoint{1.195123in}{1.226570in}}%
\pgfpathlineto{\pgfqpoint{1.237692in}{1.263175in}}%
\pgfpathlineto{\pgfqpoint{1.280260in}{1.299150in}}%
\pgfpathlineto{\pgfqpoint{1.322828in}{1.335109in}}%
\pgfpathlineto{\pgfqpoint{1.365397in}{1.371074in}}%
\pgfpathlineto{\pgfqpoint{1.407965in}{1.407008in}}%
\pgfpathlineto{\pgfqpoint{1.450533in}{1.442338in}}%
\pgfpathlineto{\pgfqpoint{1.493102in}{1.478285in}}%
\pgfpathlineto{\pgfqpoint{1.535670in}{1.513976in}}%
\pgfpathlineto{\pgfqpoint{1.578238in}{1.549451in}}%
\pgfpathlineto{\pgfqpoint{1.620807in}{1.585015in}}%
\pgfpathlineto{\pgfqpoint{1.663375in}{1.620243in}}%
\pgfpathlineto{\pgfqpoint{1.705943in}{1.655597in}}%
\pgfpathlineto{\pgfqpoint{1.748512in}{1.690901in}}%
\pgfpathlineto{\pgfqpoint{1.791080in}{1.725608in}}%
\pgfpathlineto{\pgfqpoint{1.833648in}{1.759920in}}%
\pgfpathlineto{\pgfqpoint{1.876217in}{1.794232in}}%
\pgfpathlineto{\pgfqpoint{1.918785in}{1.828554in}}%
\pgfpathlineto{\pgfqpoint{1.961353in}{1.863363in}}%
\pgfpathlineto{\pgfqpoint{2.003922in}{1.898356in}}%
\pgfpathlineto{\pgfqpoint{2.046490in}{1.933482in}}%
\pgfpathlineto{\pgfqpoint{2.089058in}{1.968709in}}%
\pgfpathlineto{\pgfqpoint{2.131627in}{2.003942in}}%
\pgfpathlineto{\pgfqpoint{2.174195in}{2.038609in}}%
\pgfpathlineto{\pgfqpoint{2.216763in}{2.073143in}}%
\pgfpathlineto{\pgfqpoint{2.259332in}{2.107523in}}%
\pgfpathlineto{\pgfqpoint{2.301900in}{2.141934in}}%
\pgfpathlineto{\pgfqpoint{2.344468in}{2.176325in}}%
\pgfpathlineto{\pgfqpoint{2.387037in}{2.211049in}}%
\pgfpathlineto{\pgfqpoint{2.429605in}{2.245953in}}%
\pgfpathlineto{\pgfqpoint{2.472173in}{2.280498in}}%
\pgfpathlineto{\pgfqpoint{2.514742in}{2.314507in}}%
\pgfpathlineto{\pgfqpoint{2.557310in}{2.348516in}}%
\pgfpathlineto{\pgfqpoint{2.599878in}{2.382670in}}%
\pgfpathlineto{\pgfqpoint{2.642447in}{2.417051in}}%
\pgfpathlineto{\pgfqpoint{2.685015in}{2.451438in}}%
\pgfpathlineto{\pgfqpoint{2.727583in}{2.485832in}}%
\pgfpathlineto{\pgfqpoint{2.770152in}{2.520226in}}%
\pgfpathlineto{\pgfqpoint{2.812720in}{2.554620in}}%
\pgfpathlineto{\pgfqpoint{2.855288in}{2.589014in}}%
\pgfpathlineto{\pgfqpoint{2.897857in}{2.623408in}}%
\pgfpathlineto{\pgfqpoint{2.940425in}{2.657908in}}%
\pgfpathlineto{\pgfqpoint{2.982993in}{2.692429in}}%
\pgfpathlineto{\pgfqpoint{3.025562in}{2.726951in}}%
\pgfpathlineto{\pgfqpoint{3.068130in}{2.761473in}}%
\pgfpathlineto{\pgfqpoint{3.110698in}{2.795995in}}%
\pgfpathlineto{\pgfqpoint{3.153267in}{2.830516in}}%
\pgfpathlineto{\pgfqpoint{3.195835in}{2.865038in}}%
\pgfpathlineto{\pgfqpoint{3.238403in}{2.899560in}}%
\pgfpathlineto{\pgfqpoint{3.280972in}{2.933978in}}%
\pgfpathlineto{\pgfqpoint{3.323540in}{2.968345in}}%
\pgfpathlineto{\pgfqpoint{3.366108in}{3.002712in}}%
\pgfpathlineto{\pgfqpoint{3.408677in}{3.037080in}}%
\pgfpathlineto{\pgfqpoint{3.451245in}{3.071447in}}%
\pgfpathlineto{\pgfqpoint{3.493813in}{3.105814in}}%
\pgfpathlineto{\pgfqpoint{3.536382in}{3.140181in}}%
\pgfpathlineto{\pgfqpoint{3.578950in}{3.174548in}}%
\pgfpathlineto{\pgfqpoint{3.621518in}{3.208920in}}%
\pgfpathlineto{\pgfqpoint{3.664087in}{3.243292in}}%
\pgfpathlineto{\pgfqpoint{3.706655in}{3.277664in}}%
\pgfpathlineto{\pgfqpoint{3.749223in}{3.312036in}}%
\pgfpathlineto{\pgfqpoint{3.791792in}{3.346588in}}%
\pgfpathlineto{\pgfqpoint{3.834360in}{3.381169in}}%
\pgfpathlineto{\pgfqpoint{3.876928in}{3.415750in}}%
\pgfpathlineto{\pgfqpoint{3.919497in}{3.450330in}}%
\pgfpathlineto{\pgfqpoint{3.962065in}{3.484911in}}%
\pgfpathlineto{\pgfqpoint{4.004633in}{3.519492in}}%
\pgfpathlineto{\pgfqpoint{4.047202in}{3.554058in}}%
\pgfpathlineto{\pgfqpoint{4.089770in}{3.588465in}}%
\pgfpathlineto{\pgfqpoint{4.132338in}{3.622871in}}%
\pgfpathlineto{\pgfqpoint{4.174907in}{3.657277in}}%
\pgfpathlineto{\pgfqpoint{4.217475in}{3.691684in}}%
\pgfpathlineto{\pgfqpoint{4.260043in}{3.726090in}}%
\pgfpathlineto{\pgfqpoint{4.302612in}{3.760497in}}%
\pgfpathlineto{\pgfqpoint{4.345180in}{3.794903in}}%
\pgfpathlineto{\pgfqpoint{4.387748in}{3.829309in}}%
\pgfpathlineto{\pgfqpoint{4.430317in}{3.863710in}}%
\pgfpathlineto{\pgfqpoint{4.472885in}{3.898075in}}%
\pgfpathlineto{\pgfqpoint{4.515453in}{3.932440in}}%
\pgfpathlineto{\pgfqpoint{4.558022in}{3.966805in}}%
\pgfpathlineto{\pgfqpoint{4.600590in}{4.001170in}}%
\pgfpathlineto{\pgfqpoint{4.643158in}{4.035535in}}%
\pgfpathlineto{\pgfqpoint{4.685727in}{4.069900in}}%
\pgfpathlineto{\pgfqpoint{4.728295in}{4.104264in}}%
\pgfpathlineto{\pgfqpoint{4.770863in}{4.138629in}}%
\pgfpathlineto{\pgfqpoint{4.813432in}{4.172994in}}%
\pgfpathlineto{\pgfqpoint{4.856000in}{4.207359in}}%
\pgfpathlineto{\pgfqpoint{4.856000in}{4.492942in}}%
\pgfpathlineto{\pgfqpoint{4.856000in}{4.492942in}}%
\pgfpathlineto{\pgfqpoint{4.813432in}{4.455103in}}%
\pgfpathlineto{\pgfqpoint{4.770863in}{4.417264in}}%
\pgfpathlineto{\pgfqpoint{4.728295in}{4.379483in}}%
\pgfpathlineto{\pgfqpoint{4.685727in}{4.341846in}}%
\pgfpathlineto{\pgfqpoint{4.643158in}{4.304210in}}%
\pgfpathlineto{\pgfqpoint{4.600590in}{4.266573in}}%
\pgfpathlineto{\pgfqpoint{4.558022in}{4.228936in}}%
\pgfpathlineto{\pgfqpoint{4.515453in}{4.191300in}}%
\pgfpathlineto{\pgfqpoint{4.472885in}{4.153663in}}%
\pgfpathlineto{\pgfqpoint{4.430317in}{4.116026in}}%
\pgfpathlineto{\pgfqpoint{4.387748in}{4.078380in}}%
\pgfpathlineto{\pgfqpoint{4.345180in}{4.040731in}}%
\pgfpathlineto{\pgfqpoint{4.302612in}{4.003082in}}%
\pgfpathlineto{\pgfqpoint{4.260043in}{3.965428in}}%
\pgfpathlineto{\pgfqpoint{4.217475in}{3.927398in}}%
\pgfpathlineto{\pgfqpoint{4.174907in}{3.889369in}}%
\pgfpathlineto{\pgfqpoint{4.132338in}{3.851339in}}%
\pgfpathlineto{\pgfqpoint{4.089770in}{3.813309in}}%
\pgfpathlineto{\pgfqpoint{4.047202in}{3.775610in}}%
\pgfpathlineto{\pgfqpoint{4.004633in}{3.737974in}}%
\pgfpathlineto{\pgfqpoint{3.962065in}{3.700338in}}%
\pgfpathlineto{\pgfqpoint{3.919497in}{3.662702in}}%
\pgfpathlineto{\pgfqpoint{3.876928in}{3.625066in}}%
\pgfpathlineto{\pgfqpoint{3.834360in}{3.587431in}}%
\pgfpathlineto{\pgfqpoint{3.791792in}{3.549795in}}%
\pgfpathlineto{\pgfqpoint{3.749223in}{3.512159in}}%
\pgfpathlineto{\pgfqpoint{3.706655in}{3.474523in}}%
\pgfpathlineto{\pgfqpoint{3.664087in}{3.436880in}}%
\pgfpathlineto{\pgfqpoint{3.621518in}{3.399233in}}%
\pgfpathlineto{\pgfqpoint{3.578950in}{3.361585in}}%
\pgfpathlineto{\pgfqpoint{3.536382in}{3.323859in}}%
\pgfpathlineto{\pgfqpoint{3.493813in}{3.285570in}}%
\pgfpathlineto{\pgfqpoint{3.451245in}{3.247157in}}%
\pgfpathlineto{\pgfqpoint{3.408677in}{3.209126in}}%
\pgfpathlineto{\pgfqpoint{3.366108in}{3.171402in}}%
\pgfpathlineto{\pgfqpoint{3.323540in}{3.133307in}}%
\pgfpathlineto{\pgfqpoint{3.280972in}{3.095914in}}%
\pgfpathlineto{\pgfqpoint{3.238403in}{3.058641in}}%
\pgfpathlineto{\pgfqpoint{3.195835in}{3.021365in}}%
\pgfpathlineto{\pgfqpoint{3.153267in}{2.984076in}}%
\pgfpathlineto{\pgfqpoint{3.110698in}{2.946790in}}%
\pgfpathlineto{\pgfqpoint{3.068130in}{2.909516in}}%
\pgfpathlineto{\pgfqpoint{3.025562in}{2.872242in}}%
\pgfpathlineto{\pgfqpoint{2.982993in}{2.834570in}}%
\pgfpathlineto{\pgfqpoint{2.940425in}{2.796458in}}%
\pgfpathlineto{\pgfqpoint{2.897857in}{2.758345in}}%
\pgfpathlineto{\pgfqpoint{2.855288in}{2.720943in}}%
\pgfpathlineto{\pgfqpoint{2.812720in}{2.683715in}}%
\pgfpathlineto{\pgfqpoint{2.770152in}{2.646486in}}%
\pgfpathlineto{\pgfqpoint{2.727583in}{2.608862in}}%
\pgfpathlineto{\pgfqpoint{2.685015in}{2.571232in}}%
\pgfpathlineto{\pgfqpoint{2.642447in}{2.533611in}}%
\pgfpathlineto{\pgfqpoint{2.599878in}{2.496119in}}%
\pgfpathlineto{\pgfqpoint{2.557310in}{2.458912in}}%
\pgfpathlineto{\pgfqpoint{2.514742in}{2.421702in}}%
\pgfpathlineto{\pgfqpoint{2.472173in}{2.384492in}}%
\pgfpathlineto{\pgfqpoint{2.429605in}{2.346695in}}%
\pgfpathlineto{\pgfqpoint{2.387037in}{2.309156in}}%
\pgfpathlineto{\pgfqpoint{2.344468in}{2.271685in}}%
\pgfpathlineto{\pgfqpoint{2.301900in}{2.234109in}}%
\pgfpathlineto{\pgfqpoint{2.259332in}{2.196759in}}%
\pgfpathlineto{\pgfqpoint{2.216763in}{2.159409in}}%
\pgfpathlineto{\pgfqpoint{2.174195in}{2.122206in}}%
\pgfpathlineto{\pgfqpoint{2.131627in}{2.085254in}}%
\pgfpathlineto{\pgfqpoint{2.089058in}{2.047819in}}%
\pgfpathlineto{\pgfqpoint{2.046490in}{2.009993in}}%
\pgfpathlineto{\pgfqpoint{2.003922in}{1.972289in}}%
\pgfpathlineto{\pgfqpoint{1.961353in}{1.934907in}}%
\pgfpathlineto{\pgfqpoint{1.918785in}{1.897578in}}%
\pgfpathlineto{\pgfqpoint{1.876217in}{1.860338in}}%
\pgfpathlineto{\pgfqpoint{1.833648in}{1.823057in}}%
\pgfpathlineto{\pgfqpoint{1.791080in}{1.785775in}}%
\pgfpathlineto{\pgfqpoint{1.748512in}{1.748495in}}%
\pgfpathlineto{\pgfqpoint{1.705943in}{1.710790in}}%
\pgfpathlineto{\pgfqpoint{1.663375in}{1.673501in}}%
\pgfpathlineto{\pgfqpoint{1.620807in}{1.636469in}}%
\pgfpathlineto{\pgfqpoint{1.578238in}{1.599551in}}%
\pgfpathlineto{\pgfqpoint{1.535670in}{1.563301in}}%
\pgfpathlineto{\pgfqpoint{1.493102in}{1.526377in}}%
\pgfpathlineto{\pgfqpoint{1.450533in}{1.489337in}}%
\pgfpathlineto{\pgfqpoint{1.407965in}{1.452318in}}%
\pgfpathlineto{\pgfqpoint{1.365397in}{1.415874in}}%
\pgfpathlineto{\pgfqpoint{1.322828in}{1.379553in}}%
\pgfpathlineto{\pgfqpoint{1.280260in}{1.342438in}}%
\pgfpathlineto{\pgfqpoint{1.237692in}{1.305722in}}%
\pgfpathlineto{\pgfqpoint{1.195123in}{1.268955in}}%
\pgfpathlineto{\pgfqpoint{1.152555in}{1.232236in}}%
\pgfpathlineto{\pgfqpoint{1.109987in}{1.196375in}}%
\pgfpathlineto{\pgfqpoint{1.067418in}{1.160061in}}%
\pgfpathlineto{\pgfqpoint{1.024850in}{1.124584in}}%
\pgfpathlineto{\pgfqpoint{0.982282in}{1.088728in}}%
\pgfpathlineto{\pgfqpoint{0.939713in}{1.053183in}}%
\pgfpathlineto{\pgfqpoint{0.897145in}{1.018055in}}%
\pgfpathlineto{\pgfqpoint{0.854577in}{0.982846in}}%
\pgfpathlineto{\pgfqpoint{0.812008in}{0.947664in}}%
\pgfpathlineto{\pgfqpoint{0.769440in}{0.912483in}}%
\pgfpathlineto{\pgfqpoint{0.726872in}{0.877099in}}%
\pgfpathlineto{\pgfqpoint{0.684303in}{0.842159in}}%
\pgfpathlineto{\pgfqpoint{0.641735in}{0.806991in}}%
\pgfpathclose%
\pgfusepath{stroke,fill}%
}%
\begin{pgfscope}%
\pgfsys@transformshift{0.000000in}{0.000000in}%
\pgfsys@useobject{currentmarker}{}%
\end{pgfscope}%
\end{pgfscope}%
\begin{pgfscope}%
\pgfpathrectangle{\pgfqpoint{0.641735in}{0.555904in}}{\pgfqpoint{4.214265in}{4.124515in}}%
\pgfusepath{clip}%
\pgfsetroundcap%
\pgfsetroundjoin%
\pgfsetlinewidth{1.806750pt}%
\definecolor{currentstroke}{rgb}{0.400000,0.760784,0.647059}%
\pgfsetstrokecolor{currentstroke}%
\pgfsetdash{}{0pt}%
\pgfpathmoveto{\pgfqpoint{0.641735in}{0.778800in}}%
\pgfpathlineto{\pgfqpoint{0.684303in}{0.814983in}}%
\pgfpathlineto{\pgfqpoint{0.726872in}{0.851167in}}%
\pgfpathlineto{\pgfqpoint{0.769440in}{0.887350in}}%
\pgfpathlineto{\pgfqpoint{0.812008in}{0.923533in}}%
\pgfpathlineto{\pgfqpoint{0.854577in}{0.959716in}}%
\pgfpathlineto{\pgfqpoint{0.897145in}{0.995899in}}%
\pgfpathlineto{\pgfqpoint{0.939713in}{1.032083in}}%
\pgfpathlineto{\pgfqpoint{0.982282in}{1.068266in}}%
\pgfpathlineto{\pgfqpoint{1.024850in}{1.104449in}}%
\pgfpathlineto{\pgfqpoint{1.067418in}{1.140632in}}%
\pgfpathlineto{\pgfqpoint{1.109987in}{1.176816in}}%
\pgfpathlineto{\pgfqpoint{1.152555in}{1.212999in}}%
\pgfpathlineto{\pgfqpoint{1.195123in}{1.249182in}}%
\pgfpathlineto{\pgfqpoint{1.237692in}{1.285365in}}%
\pgfpathlineto{\pgfqpoint{1.280260in}{1.321549in}}%
\pgfpathlineto{\pgfqpoint{1.322828in}{1.357732in}}%
\pgfpathlineto{\pgfqpoint{1.365397in}{1.393915in}}%
\pgfpathlineto{\pgfqpoint{1.407965in}{1.430098in}}%
\pgfpathlineto{\pgfqpoint{1.450533in}{1.466282in}}%
\pgfpathlineto{\pgfqpoint{1.493102in}{1.502465in}}%
\pgfpathlineto{\pgfqpoint{1.535670in}{1.538648in}}%
\pgfpathlineto{\pgfqpoint{1.578238in}{1.574831in}}%
\pgfpathlineto{\pgfqpoint{1.620807in}{1.611015in}}%
\pgfpathlineto{\pgfqpoint{1.663375in}{1.647198in}}%
\pgfpathlineto{\pgfqpoint{1.705943in}{1.683381in}}%
\pgfpathlineto{\pgfqpoint{1.748512in}{1.719564in}}%
\pgfpathlineto{\pgfqpoint{1.791080in}{1.755747in}}%
\pgfpathlineto{\pgfqpoint{1.833648in}{1.791931in}}%
\pgfpathlineto{\pgfqpoint{1.876217in}{1.828114in}}%
\pgfpathlineto{\pgfqpoint{1.918785in}{1.864297in}}%
\pgfpathlineto{\pgfqpoint{1.961353in}{1.900480in}}%
\pgfpathlineto{\pgfqpoint{2.003922in}{1.936664in}}%
\pgfpathlineto{\pgfqpoint{2.046490in}{1.972847in}}%
\pgfpathlineto{\pgfqpoint{2.089058in}{2.009030in}}%
\pgfpathlineto{\pgfqpoint{2.131627in}{2.045213in}}%
\pgfpathlineto{\pgfqpoint{2.174195in}{2.081397in}}%
\pgfpathlineto{\pgfqpoint{2.216763in}{2.117580in}}%
\pgfpathlineto{\pgfqpoint{2.259332in}{2.153763in}}%
\pgfpathlineto{\pgfqpoint{2.301900in}{2.189946in}}%
\pgfpathlineto{\pgfqpoint{2.344468in}{2.226130in}}%
\pgfpathlineto{\pgfqpoint{2.387037in}{2.262313in}}%
\pgfpathlineto{\pgfqpoint{2.429605in}{2.298496in}}%
\pgfpathlineto{\pgfqpoint{2.472173in}{2.334679in}}%
\pgfpathlineto{\pgfqpoint{2.514742in}{2.370863in}}%
\pgfpathlineto{\pgfqpoint{2.557310in}{2.407046in}}%
\pgfpathlineto{\pgfqpoint{2.599878in}{2.443229in}}%
\pgfpathlineto{\pgfqpoint{2.642447in}{2.479412in}}%
\pgfpathlineto{\pgfqpoint{2.685015in}{2.515595in}}%
\pgfpathlineto{\pgfqpoint{2.727583in}{2.551779in}}%
\pgfpathlineto{\pgfqpoint{2.770152in}{2.587962in}}%
\pgfpathlineto{\pgfqpoint{2.812720in}{2.624145in}}%
\pgfpathlineto{\pgfqpoint{2.855288in}{2.660328in}}%
\pgfpathlineto{\pgfqpoint{2.897857in}{2.696512in}}%
\pgfpathlineto{\pgfqpoint{2.940425in}{2.732695in}}%
\pgfpathlineto{\pgfqpoint{2.982993in}{2.768878in}}%
\pgfpathlineto{\pgfqpoint{3.025562in}{2.805061in}}%
\pgfpathlineto{\pgfqpoint{3.068130in}{2.841245in}}%
\pgfpathlineto{\pgfqpoint{3.110698in}{2.877428in}}%
\pgfpathlineto{\pgfqpoint{3.153267in}{2.913611in}}%
\pgfpathlineto{\pgfqpoint{3.195835in}{2.949794in}}%
\pgfpathlineto{\pgfqpoint{3.238403in}{2.985978in}}%
\pgfpathlineto{\pgfqpoint{3.280972in}{3.022161in}}%
\pgfpathlineto{\pgfqpoint{3.323540in}{3.058344in}}%
\pgfpathlineto{\pgfqpoint{3.366108in}{3.094527in}}%
\pgfpathlineto{\pgfqpoint{3.408677in}{3.130711in}}%
\pgfpathlineto{\pgfqpoint{3.451245in}{3.166894in}}%
\pgfpathlineto{\pgfqpoint{3.493813in}{3.203077in}}%
\pgfpathlineto{\pgfqpoint{3.536382in}{3.239260in}}%
\pgfpathlineto{\pgfqpoint{3.578950in}{3.275443in}}%
\pgfpathlineto{\pgfqpoint{3.621518in}{3.311627in}}%
\pgfpathlineto{\pgfqpoint{3.664087in}{3.347810in}}%
\pgfpathlineto{\pgfqpoint{3.706655in}{3.383993in}}%
\pgfpathlineto{\pgfqpoint{3.749223in}{3.420176in}}%
\pgfpathlineto{\pgfqpoint{3.791792in}{3.456360in}}%
\pgfpathlineto{\pgfqpoint{3.834360in}{3.492543in}}%
\pgfpathlineto{\pgfqpoint{3.876928in}{3.528726in}}%
\pgfpathlineto{\pgfqpoint{3.919497in}{3.564909in}}%
\pgfpathlineto{\pgfqpoint{3.962065in}{3.601093in}}%
\pgfpathlineto{\pgfqpoint{4.004633in}{3.637276in}}%
\pgfpathlineto{\pgfqpoint{4.047202in}{3.673459in}}%
\pgfpathlineto{\pgfqpoint{4.089770in}{3.709642in}}%
\pgfpathlineto{\pgfqpoint{4.132338in}{3.745826in}}%
\pgfpathlineto{\pgfqpoint{4.174907in}{3.782009in}}%
\pgfpathlineto{\pgfqpoint{4.217475in}{3.818192in}}%
\pgfpathlineto{\pgfqpoint{4.260043in}{3.854375in}}%
\pgfpathlineto{\pgfqpoint{4.302612in}{3.890559in}}%
\pgfpathlineto{\pgfqpoint{4.345180in}{3.926742in}}%
\pgfpathlineto{\pgfqpoint{4.387748in}{3.962925in}}%
\pgfpathlineto{\pgfqpoint{4.430317in}{3.999108in}}%
\pgfpathlineto{\pgfqpoint{4.472885in}{4.035291in}}%
\pgfpathlineto{\pgfqpoint{4.515453in}{4.071475in}}%
\pgfpathlineto{\pgfqpoint{4.558022in}{4.107658in}}%
\pgfpathlineto{\pgfqpoint{4.600590in}{4.143841in}}%
\pgfpathlineto{\pgfqpoint{4.643158in}{4.180024in}}%
\pgfpathlineto{\pgfqpoint{4.685727in}{4.216208in}}%
\pgfpathlineto{\pgfqpoint{4.728295in}{4.252391in}}%
\pgfpathlineto{\pgfqpoint{4.770863in}{4.288574in}}%
\pgfpathlineto{\pgfqpoint{4.813432in}{4.324757in}}%
\pgfpathlineto{\pgfqpoint{4.856000in}{4.360941in}}%
\pgfusepath{stroke}%
\end{pgfscope}%
\begin{pgfscope}%
\pgfsetrectcap%
\pgfsetmiterjoin%
\pgfsetlinewidth{1.003750pt}%
\definecolor{currentstroke}{rgb}{0.800000,0.800000,0.800000}%
\pgfsetstrokecolor{currentstroke}%
\pgfsetdash{}{0pt}%
\pgfpathmoveto{\pgfqpoint{0.641735in}{0.555904in}}%
\pgfpathlineto{\pgfqpoint{0.641735in}{4.680420in}}%
\pgfusepath{stroke}%
\end{pgfscope}%
\begin{pgfscope}%
\pgfsetrectcap%
\pgfsetmiterjoin%
\pgfsetlinewidth{1.003750pt}%
\definecolor{currentstroke}{rgb}{0.800000,0.800000,0.800000}%
\pgfsetstrokecolor{currentstroke}%
\pgfsetdash{}{0pt}%
\pgfpathmoveto{\pgfqpoint{0.641735in}{0.555904in}}%
\pgfpathlineto{\pgfqpoint{4.856000in}{0.555904in}}%
\pgfusepath{stroke}%
\end{pgfscope}%
\begin{pgfscope}%
\definecolor{textcolor}{rgb}{0.150000,0.150000,0.150000}%
\pgfsetstrokecolor{textcolor}%
\pgfsetfillcolor{textcolor}%
\pgftext[x=2.748867in,y=4.763753in,,base]{\color{textcolor}\sffamily\fontsize{9.600000}{11.520000}\selectfont Linear growth in game size}%
\end{pgfscope}%
\end{pgfpicture}%
\makeatother%
\endgroup%

%% file: related.tex
\section{Discussion and Related work}\label{sec:related}

\subsection{Control Improvisation in the Literature}

In this section, we briefly compare ERCI with other forms of control
improvisation. Firstly, we observe that general Control Improvisation
has been proposed in stochastic environments for lane
changing~\cite{DBLP:conf/cdc/GeM18} and imitating power usage in
households~\cite{DBLP:conf/iotdi/AkkayaFVDLS16}. However, in those
both settings, the randomness constraint is phrased as an upper-bound
on the probability of indefinitely-long paths. Consequently, those
randomness constraints are trivially satisfied.  In comparison, we consider the synthesis
of policies that necessarily randomize in presence of stochastic
behavior in the environment. The closest prior work is to ours
is Reactive Control Improvisation (RCI) for (deterministic) 2-player games~\cite{DBLP:conf/cav/FremontS18}. As in ERCI, RCI features
 three kinds of constraints; hard, soft, and randomness. As in ERCI,
RCI can be preprocessed resulting in the following core problem.
\begin{mdframed}[nobreak=true]
  \textbf{The Core RCI Problem}: Given a finite acyclic
  (deterministic) SG $\sg$, with terminal states, $\target$ and $\sink$,
  and thresholds $\scthreshold \in (0,1)$ and
  $\randomness \in [0,\infty)$, find a $\pOne$-policy $\pOneSched$
  such that for every $\pTwo$-policy $\pTwoSched$
  \begin{enumerate}
  \item (\emph{soft constraint)}
    $\Pr(\last{\xi} = \target \mid \sched) \geq \scthreshold$,
  \item (\emph{randomness})
    $\max_\xi{\Pr(\path \mid \sched)} \leq \ppthreshold$,
   \end{enumerate}
   where $\sched = \langle \pOneSched, \pTwoSched \rangle$.
 \end{mdframed}
 While RCI is only applied to deterministic SGs in
 \cite{DBLP:conf/cav/FremontS18}, there is nothing in the definition
 that prevents its application to the general class of
 SGs.\footnote{However, this does not mean that the algorithm to
   compute a solution carries over to the general case} We observe
 that then, the only difference between ERCI and RCI is that we use
 causal entropy rather than an upper bound on the probability of a
 path to enforce randomness.  Below we address two problems with
 bounding the maximum probability of a trace.

 First, RCI fails to account for causality when measuring
 randomness. In deterministic systems, for which RCI was conceived,
 this distinction is unnecessary, but stochastic systems must deal
 with counter-factuals. In practice, RCI encodes an agent model that
 is systematically overly optimistic regarding the outcomes of
 dynamics transitions~\cite{DBLP:journals/corr/abs-1805-00909}. This
 results in policies with worse performance given a fixed randomness
 target.  In the context of our motivating drone example, applying RCI
 thus results in a policy that is both quantitatively and
 qualitatively less random than the ERCI.

 Second, RCI fails to enforce randomization if there exists \emph{any} path
 with sufficiently high probability. The next (pathological) example
 illustrates.
\begin{figure}
\centering
\scalebox{0.8}{
\begin{tikzpicture}
        \node[sstate, initial, initial text=] (s0) {$s_0$};
  	\node[actnode,right=of s0, xshift=-2em] (e0) {};
        
	\node[sstate,right=of s0, yshift=2.5em] (s2) {$s_2$};
        \node[sstate,above=0.4cm of s2] (s1) {$s_1$};
	\node[below=0.1cm of s2] (sdots) {$\vdots$};
	\node[sstate,below=0.1cm of sdots] (sn) {$s_n$};
        
	\node[sstate,right=of s1] (t1) {$t_1$};
	\node[right=of t1] (tdots) {$\hdots$};
	\node[sstate,right=of tdots] (tn) {$t_m$};
	\node[sstate, right=2.2cm of e0] (u) {$u$};
	\node[sstate, right=of u, yshift=1em] (v1) {$v_1$};
	\node[sstate, right=of u, yshift=-1em] (v2) {$v_2$};
	\node[right=0.5cm of v1] {$\hdots$};
	\node[right=0.5cm of v2] {$\hdots$};

	\draw[->] (t1) -- node[elab] {$1$} (tdots);
	\draw[->] (tdots) -- node[elab] {$1$} (tn);

	\draw[->] (s0) -- node[elab] {} (e0);
        
	\draw[->] (e0) -- node[elab] {$\nicefrac{1}{n}$} (s1);
	\draw[->] (e0) -- node[elab,below,xshift=1mm] {$\nicefrac{1}{n}$} (s2);
	\draw[->] (e0) -- node[elab,below,xshift=-1mm] {$\nicefrac{1}{n}$} (sn);
	
	\draw[->] (s1) -- node[elab,above] {$1$} (t1);
	\draw[->] (s2) -- node[elab,near end,above] {$1$} (u);
	\draw[->] (sn) -- node[elab,near end,below] {$1$} (u);
	
	\draw[->] (u) -- node[actnode] {} node[elab,above, near start] {$a$}  node[elab,above, near end] {$1$} (v1);
	\draw[->] (u) -- node[actnode] {} node[elab,below, near start] {$b$} node[elab,below, near end] {$1$} (v2);

\end{tikzpicture}
}
\caption{Example Illustrating the problem with RCI in stochastic environments.\label{fig:drciOnSgs}}
\end{figure}
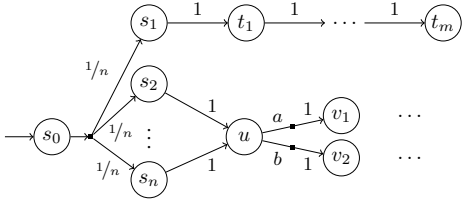

\begin{example}
  Consider the SG (actually, an MDP where we omit the $\pTwo$-states) in Fig.~\ref{fig:drciOnSgs}.
  First consider that under each scheduler, the path from $s_0$ to
  $t_m$ has probability $\nicefrac{1}{n}$. In particular, this means
  that a feasible RCI instance (applied to an SG) must have
  $\ppthreshold \geq \nicefrac{1}{n}$. At the same time, every path in
  the SG already has probability at most $\nicefrac{1}{n}$, and thus,
  every scheduler that satisfies the randomness constraint for
  $\delta = 1$ satisfies it for any
  $\ppthreshold \geq \nicefrac{1}{n}$. Thus, for this MDP, the RCI formulation fails to 
  enforce any randomization in the $\pOne$-policy. By contrast, a
  causal entropy constraint from ERCI will continuously trade-off randomness
  for performance.
\end{example}

% On the other hand, for deterministic SGs, all randomization is due to
% the random behavior of $\pOne$. 

% Roughly speaking, every
% (controllable) path is then just good (i.e., ending in a target state)
% or bad (i.e., ending in a sink state). The randomization criterion
% then may just ensure that we do not select the same path with too much
% probability, which intuitively means that the $\pOne$-player must
% ensure in every step that there are a sufficient number of paths from
% the next state (no matter what $\pTwo$-player does).  As long as there
% exist a sufficient number of paths, randomizing (uniformly) over those
% paths leads to a solution.

On the other hand, one can observe that in reality, proposed
algorithms for solving RCI equally distribute probability mass across
the maximum number of paths that $\pOne$ can
guarantee~\cite{DBLP:conf/cav/FremontS18}. We remark that because (1)
causal entropy reduces to non-causal entropy in deterministic dynamics
and (2) uniform distributions maximize entropy, our proposed entropy
matching family exactly agrees with existing RCI algorithms on
deterministic SGs. Thus, we observe the following proposition.
\begin{proposition}\label{prop:conservative}
  There exists a computable function,
  \[f \colon (\ppthreshold, \sg) \mapsto \randomness,\] such that, for any
  deterministic SG, $\sg$, and performance threshold $\scthreshold$,
  there exists an $\pOne$-policy solving the RCI problem with
  threshold $\ppthreshold$ iff there exists a $\pOne$-policy solving
  the ERCI problem with threshold
  $\randomness = f(\ppthreshold, \sg)$.
\end{proposition}

\subsection{Additional Related Work}
 
Synthesis in MDPs with multiple hard and soft constraints (often over indefinite horizons) is a well-studied problem~\cite{DBLP:conf/stacs/ChatterjeeMH06,DBLP:conf/tacas/EtessamiKVY07,DBLP:conf/atva/ForejtKP12,DBLP:journals/fmsd/RandourRS17}.  In this setting, one generates deterministic policies and their convex combinations. Put differently, some degree of randomization is \emph{not an objective}, but rather a consequence. Interestingly, in \cite{DBLP:conf/tacas/DelgrangeKQR20} the optimal policies in \emph{absence} of randomization are investigated. Along similar lines, \cite{DBLP:journals/jcss/BrazdilCFK17} trades average performance for less variance, thereby implicitly trading off the average and the worst-case performance.  
The original results sparked interest in different extension to MDPs and the type of soft constraints, such as continuous MDPs \cite{DBLP:journals/csysl/HaesaertNS21} and continuous-time MDPs~\cite{DBLP:conf/cav/QuatmannJK17},  cost-bounded reachability \cite{DBLP:journals/jar/HartmannsJKQ20}, or mean-payoff properties~\cite{DBLP:journals/corr/abs-1104-3489}. 
The algorithms have also been extended towards stochastic games~\cite{DBLP:conf/mfcs/ChenFKSW13,DBLP:journals/sttt/KwiatkowskaPW18}.
Finally, notions of lexicographic multi-objective synthesis~\cite{DBLP:conf/cav/ChatterjeeKWW20} -- in which one optimizes a secondary criterion among all policies that are optimal with respect to a first criterion bare some resemblance with the algorithm we consider. 
The aforementioned algorithms have been put in a robotics context in~\cite{DBLP:journals/ijrr/LacerdaFPH19}.
Finding policies that optimize reward objectives is well-studied in the field of reinforcement learning, and has been extended to generate Pareto fronts for multiple objectives~\cite{DBLP:conf/icml/NatarajanT05,DBLP:conf/adprl/ParisiPSBR14}.

Next, our core ERCI instance can be seen as a multi-objective path problem~\cite{DBLP:conf/icra/AmigoniG05,DBLP:journals/eswa/NazarahariKD19,DBLP:conf/icml/XuTMRSM20}.
The literature on multi-object path finding differs prominently from ERCI in two aspects: they do not trade-off
randomization and performance, and they do not trade-off declarative
and formal constraints with the accompanying formal guarentees, but
are more search-based.

Another related domain is the problem of (randomly) patrolling a perimeters and
points of interest~\cite{DBLP:conf/icra/AgmonKK08,DBLP:conf/icra/AmigoniBG09,DBLP:conf/iros/PortugalPRC14}.
Closest to our work are  formalisms rooted in game-theory,  such as  \emph{Stackelberg games}~\cite{simaan1973stackelberg,DBLP:conf/atal/ParuchuriPTOK07}. Stackelberg games have been extending to Stackelberg planning~\cite{DBLP:conf/aaai/SpeicherS00K18} in which a trade-off between the cost for the defender and the attacker can be investigated.
Most related are the zero-sum~\emph{patrolling games} introduced in~\cite{DBLP:journals/ior/AlpernMP11}, which has led to numerous practical solutions~\cite{DBLP:books/daglib/0040483}. Patrolling games are explicitly games between an intruder and a defender, and there is no stochastic environment.  Adding additional objectives makes solving these problems harder~\cite{DBLP:conf/atal/Klaska0R20} and in general, the obtained policies are no longer applicable. To overcome this, a specific set of fixed objectives has been added to these games recently~\cite{DBLP:conf/atal/Klaska0R20}. 
 The large common aspect in all of this work is that optimal strategies do randomize. As in the synthesis work above, this is a consequence of the objectives rather than an objective in itself. 
 In comparison, we provide a general framework and in particular support stochastic environments.

Finally, entropy as an optimization objective for MDPs with fixed
rewards has been well studied~\cite{DBLP:journals/tac/SavasOCKT20},
particularly in the context of regularizing (robustifying) inverse and
reinforcement learning~\cite{mceThesis, DBLP:conf/icml/GeistSP19}. The
primary distinction from our work (in the MDP setting) is the
unspecified (performance/entropy) trade-off. Nevertheless, as
previously discussed, the specification varient of this literature
served as the basis for our MDP
subroutine~\cite{DBLP:conf/cav/Vazquez-Chanlatte20}.  Beyond Markov
models, the (uniform) randomization over languages in finite automata
\cite{DBLP:journals/siamcomp/HickeyC83,DBLP:conf/soda/KannanSM95} or
over propositional formulae
\cite{DBLP:journals/tcs/JerrumVV86,DBLP:journals/iandc/BellareGP00,DBLP:conf/dac/ChakrabortyMV14}
has received quite some attention, however neither of those approaches
support the notion of soft constraints or the related trade-offs.

%%% Local Variables:
%%% mode: latex
%%% TeX-master: "main"
%%% End:

%% file: conclusion.tex
\section{Conclusion}
This paper presented ERCI, a framework to control improvisation in stochastic games. Our results show that ERCI can be used to synthesize policies that besides meeting temporal logic specifications induce varying behavior, e.g., to test and certify the correctness of other robots. Future work includes applying the framework to a broader spectrum of applications and extending the theory to games with imperfect information.

%% file: proofs.tex
\section{Proofs}\label{sec:proofs}
\subsection{Convexity of ERCI solution set}
\begin{proof}[Proof Sketch Prop~\ref{prop:convex}]
  Recall that a set is convex, if it is closed under
  convex-combinations\footnotemark. Consider two points
  $\langle \scp, \rndp \rangle, \langle \scp', \rndp' \rangle \in
  \solutions$ achieved by $\pOneSched$ and $\pOneSchedPrime$
  respectively. Consider the new policy, $\pi$, defined by employing
  $\pOneSched$ with probability $q$ and $\pOneSchedPrime$ with
  probability $\bar{q} \eqdef 1 - q$.  Because each policy
  \emph{guarantees} its corresponding performance, this new policy has
  performance at least $q\cdot \scp + \bar{q}\cdot \scp'$.  Similarly,
  by viewing $\pi$ as a random variable and applying chain rule
  yields,
  \begin{equation}
    \begin{split}
      H_\tau(\sigma)
      \geq &~q \cdot H( \rv{A}^{\pOne}_{1:\tau'} \mid\mid \rv{S}_{1:\tau} \mid \pi=\pOneSched)~+\\
      &~\bar{q}  \cdot H( \rv{A}^{\pOne}_{1:\tau'} \mid\mid \rv{S}_{1:\tau} \mid \pi=\pOneSchedPrime)\\
      =&~q\cdot\rndp + \bar{q}\cdot \rndp'.
    \end{split}
  \end{equation}
  Thus, any convex combination of guaranteed points is guaranteed by
  a convex combination of the corresponding ego policies.
\end{proof}

\subsection{Completeness of Entropy Matching for SGs}

\begin{proof}[Proof Sketch of SG Completeness]
  We prove the statement by induction over the (acyclic) SG.
  First, observe that on games with only terminal nodes, completeness
  follows directly.  Next, suppose the entropy matching family is
  complete on all sub-graphs of $\sg$. To simplify our proof, observe
  that w.l.o.g., we can restrict our attention to ERCI instances
  on the Pareto front, $\langle \scthreshold,\randomness\rangle \in \pareto{\solutions}$.
  Next, for the sake of contradiction, we shall assume that no entropy
  matching policy achieves $\langle \scthreshold,\randomness\rangle$,
  but $\sched_{\pOne}^*$ does:
  \begin{align}
    &\forall \sched_\pOne \in \{\sched_{\pOne}^\rat\}_\rat~.~ x_{\sched_{\pOne}} \prec \langle \scthreshold, \randomness \rangle\label{eq:reject}\\
    &\exists \sched_\pOne^* \notin \{\sched_{\pOne}^\rat\}_\rat~.~  \langle \scthreshold, \randomness \rangle \preceq x_{\sched_{\pOne}^*}\label{eq:incomplete}.
  \end{align}
  Indeed, we may reformulate \eqref{eq:incomplete} to 
  \begin{align}
  	\exists \sched_\pOne^* \notin \{\sched_{\pOne}^\rat\}_\rat~.~  \langle \scthreshold, \randomness \rangle = x_{\sched_{\pOne}^*}\label{eq:incompleteeq}
  \end{align}
as we assumed that $\langle \scthreshold, \randomness \rangle$ is Pareto-optimal.
  
  Note that because the entropy matching family contains the maximizers and minimizers
  of entropy ($\rat = \infty$ and $\rat = 0$ resp.), and because increasing rationality monotonically decreases entropy,
  there must exist some rationality, $\rat$, such that $\sched_\pOne^\rat$ induces entropy $\randomness$:
  \begin{equation}
    \rndp_{\sched_\pOne^\rat} = \randomness = \rndp_{\sched_\pOne^*},
  \end{equation}
  where the second equality follows from \eqref{eq:incompleteeq}.
  Next, let $\sched_\pTwo^\rat$ denote the min-entropy
  $\pTwo$-policy given $\sched_\pOne^{\rat}$, i.e., the policy that minimizes entropy in $\sg[\sched_\pOne^{\rat}]$.  
  Because $\sched_\pOne^*$
  witnesses $\langle\scthreshold, \randomness\rangle$, it must be the case
  that:
  \begin{equation}\label{eq:pareto_bound_for_rat}
    \rndp_{\langle \sched_\pOne^*, \sched_\pTwo^\rat \rangle} \geq \randomness
    \hspace{2em} \text{and} \hspace{2em}
    \scp_{\langle \sched_\pOne^*, \sched_\pTwo^\rat\rangle} \geq \scthreshold
  \end{equation}
  Recalling that for MDPs, the maximum entropy policies as defined in \eqref{eq:mdp:first}--\eqref{eq:mdp:last}  are the unique maximizers of entropy (given $\scp$),
  it must be the case that:
  \begin{equation}
    \randomness = \rndp_{\langle \sched_\pOne^\rat \sched_\pTwo^\rat \rangle} \geq  \rndp_{\langle \sched_\pOne^*, \sched_\pTwo^\rat \rangle}\geq \randomness,
  \end{equation}
  and thus,
  \begin{equation}
    \rndp_{\langle \sched_\pOne^\rat \sched_\pTwo^\rat \rangle} =  \rndp_{\langle \sched_\pOne^*, \sched_\pTwo^\rat \rangle}.
  \end{equation}
  Thus, from uniqueness on MDPs, $\sched_\pOne^\rat$ and
  $\sched_\pOne^*$ must exactly match on $\sg[\sched_\pTwo^\rat]$ and
  must differ on some other subgraph.  Applying the inductive
  hypothesis, we know that the entropy matching family is complete on
  these subgraphs, and thus if $\sched_\pOne^*$ achieves a given
  $\langle\scthreshold, \randomness\rangle$ on this subgraph, there
  must be an entropy matching that does so as well. Thus,
  \begin{equation}
    x_{\sched_{\pOne}^*} \preceq x_{\sched_{\rat^*}},
  \end{equation}
  contradicting
  assumptions~\eqref{eq:reject} and \eqref{eq:incomplete}.  Thus,
  entropy matching must be complete.
\end{proof}

%%% Local Variables:
%%% mode: latex
%%% TeX-master: "main"
%%% End: